\documentclass[pmlr]{jmlr}%
\DeclareUnicodeCharacter{2212}{\textminus}

\RequirePackage{graphicx}
 \usepackage{booktabs}
\usepackage{longtable}%

 \usepackage{adjustbox}
 \usepackage{comment}
 \usepackage{subcaption}

\makeatletter
\def\set@curr@file#1{\def\@curr@file{#1}} %
\makeatother
\usepackage[load-configurations=version-1]{siunitx} %

\theorembodyfont{\upshape}
\theoremheaderfont{\scshape}
\theorempostheader{:}
\theoremsep{\newline}

\jmlrproceedings{PMLR}{Proceedings of Machine Learning Research}
\jmlrvolume{340}
\jmlryear{2026}
\jmlrworkshop{Machine Learning for Healthcare}

\title[Learning Under Treatment-Induced Label Indeterminacy]{Learning Under Treatment-Induced Label Indeterminacy with Expert Annotations of Counterfactual Outcomes: \\ A Case Study in Neurological Prognostication}

\author{\Name{Xiaobin Shen} %
       \Email{xiaobins@andrew.cmu.edu}\\ 
       \addr Heinz College of Information Systems and Public Policy\\
       Carnegie Mellon University\\ \vspace{-0.8em}
       \\ 
       \Name{Chloe Y.H. Huang} %
       \Email{yihanhua@andrew.cmu.edu}\\ 
       \addr Heinz College of Information Systems and Public Policy\\
       Carnegie Mellon University\\ \vspace{-0.8em}
       \\ 
       \Name{Jonathan Elmer} %
       \Email{elmerjp@upmc.edu}\\ 
       \addr Department of Emergency Medicine\\
       University of Pittsburgh\\ \vspace{-0.8em}
       \\
       \Name{George H. Chen} %
       \Email{georgechen@cmu.edu}\\ 
       \addr Heinz College of Information Systems and Public Policy\\
       Carnegie Mellon University\\
       }

\begin{document}

\maketitle
\vspace{-2.9em}
\begin{abstract}
    Clinical prediction models are often developed as if the outcome of interest were cleanly observed for every patient. This assumption fails when treatment decisions make the clinically relevant outcome permanently unobservable. As a case study of this problem, we consider post-cardiac-arrest neurological prognostication using a cohort of 2,497 patients, including 1,429 patients whose outcomes were rendered indeterminate by treatment decisions
    (e.g., withdrawing or limiting life-sustaining therapies, which immediately led to death, so we do not know what would have happened otherwise). 
    These patients with indeterminate outcomes were reviewed by independent clinical experts, who provided their guesses of counterfactual outcomes about what would have happened to the patients. We refer to these patients as \emph{uncertain} cases. We also have patients for whom we observe their clinically relevant outcomes (e.g., regaining consciousness); we refer to these patients as \emph{certain} cases.
    We propose a framework for evaluating prediction models that explicitly splits the evaluation between certain and uncertain cases. Here, we cannot easily evaluate both types of cases in a uniform manner as the available target labels differ (we have known outcomes for certain cases, and guesses of counterfactual outcomes for uncertain cases).
    We then propose a simple prediction model that uses target labels from both certain and uncertain cases in a manner that allows us to trade off between them. 
    \textcolor{black}{Across the proposed neural model and a collection of tabular baselines, models with similar certain-case AUROC can nevertheless differ substantially in both certain-case Brier score and their probability estimates for uncertain cases.}
    Improving alignment with target labels of uncertain cases for our proposed model generally comes at the cost of worse accuracy on certain cases, highlighting an explicit tradeoff that standard evaluation (focusing only on certain cases) conceals. These results show that when treatment decisions determine whether clinically meaningful outcomes remain observable, conventional evaluation metrics can miss important failure modes in the very patients for whom prognostic support matters most. 
    Code is available at \url{https://github.com/xiaobin-xs/learning-under-label-indeterminacy}. 
  
\end{abstract}

\section{Introduction}

Accurate prognostication after cardiac arrest is both clinically urgent and methodologically challenging. Among patients who survive the initial arrest and are admitted to the hospital, many remain comatose and require ongoing life-sustaining therapies, yet their eventual neurological recovery is highly uncertain~\citep{nih2025, martin2024}. In this setting, clinicians and families must decide whether continued aggressive treatment is likely to lead to meaningful recovery. These decisions are consequential, time-sensitive, and often vary across physicians and hospitals~\citep{steinberg2022, cobler2025}.

The setting just described creates a fundamental problem for prediction models. Supervised learning typically assumes that the target label is a meaningful observation of the outcome of interest. After cardiac arrest, however, this assumption can fail precisely in the cases where prognostic support matters most. 
\textcolor{black}{For example, when life-sustaining therapies are withdrawn before neurological recovery can be observed, the patient’s subsequent death does not reveal whether recovery might have occurred had treatment continued. A related problem arises when multisystem organ failure, rearrest, or another competing terminal event prevents continued recovery-directed treatment or causes the patient to die before neurological recovery can be assessed. Although these pathways differ clinically, both prevent observation of the target outcome: neurological recovery under continued treatment.} This problem is referred to as \emph{treatment-induced label indeterminacy} \citep{schoeffer2025},
and is related to the selective labels problem~\citep{lakkaraju2017}, where only a subset of instances (patients in our setting) have observed outcomes, and whether an observed outcome is usable for the prediction depends on historical decisions, such as WLST. %
\textcolor{black}{For ease of exposition, throughout this paper we use the term \textbf{limiting action} as shorthand for a treatment-limiting process. This includes both explicit treatment-limitation decisions, such as withdrawal of life-sustaining therapies, and competing terminal clinical events, such as multisystem organ failure or rearrest, that prevent the target continued-treatment recovery trajectory from being observed.}

\begin{figure}[t]
    \centering
    \includegraphics[width=0.9\textwidth]{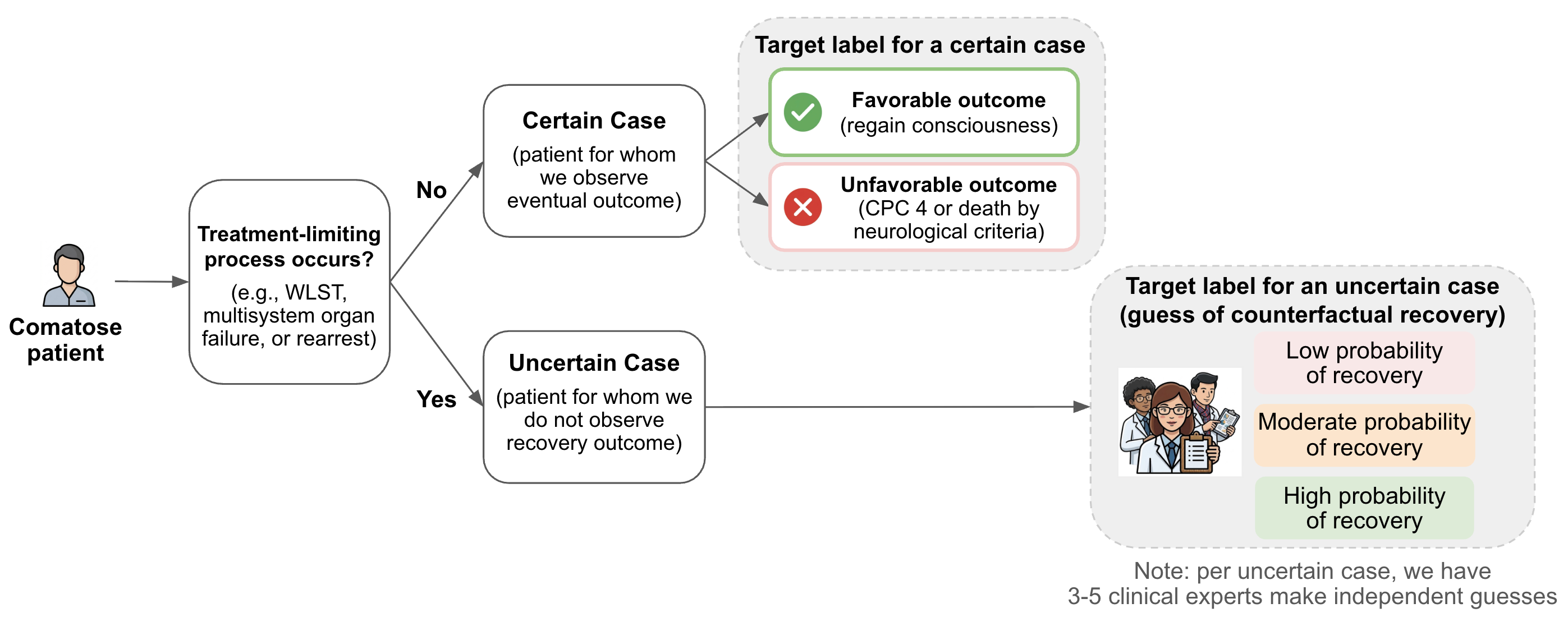}
    \vspace{-0.5em}
     \caption{\textbf{Target labels in the split evaluation setting.} %
     Each patient either experiences a \textbf{treatment-limiting process} or not. 
     \textcolor{black}{A treatment-limiting process may be either an explicit treatment-limitation decision, such as withdrawal of life-sustaining therapy (WLST), or a competing terminal clinical event, such as multisystem organ failure or rearrest, that prevents neurological recovery under continued treatment from being observed. In the WLST pathway, treatment withdrawal directly precedes death; in the competing-event pathway, another terminal condition interrupts or constrains the recovery-directed clinical trajectory.}
     We refer to patients who experience a treatment-limiting process as \textbf{uncertain} cases because their neurological outcomes under continued treatment are not observed. Each uncertain case is independently reviewed by multiple clinical experts, who provide structured assessments of counterfactual recovery, such as the estimated probability of regaining consciousness and surviving to hospital discharge. Patients who do not experience a limiting action are treated as \textbf{certain} cases, for whom the relevant clinical outcome is observed. 
     Our paper crucially uses both types of target labels to develop an evaluation framework and also a prediction model.
     } %
    \label{fig:label-indeterminacy-overview}
\end{figure}

A key feature of our study is access to expert assessments collected in prior peer-reviewed clinical research \citep{elmer2025recovery}. For patients whose recovery under continued treatment could not be observed due to the limiting action, multiple experts independently estimated the probability that the patient would have regained consciousness and survived to hospital discharge had life-sustaining therapies continued.
We use these assessments as imperfect reference signals about expert belief, not as observed outcomes or ground-truth counterfactual labels. Each uncertain case in our analytic cohort received 3–5 assessments.
A full diagram of the target labels that we obtain is shown in Figure~\ref{fig:label-indeterminacy-overview}. Note that we refer to patients who undergo a limiting action as \textbf{uncertain} cases in the sense that we are not certain about what would have happened to them (had they not experienced a limiting action): their target labels are guesses of counterfactual outcomes. In contrast, we refer to the patients that do not undergo a limiting action as \textbf{certain} cases in that we directly observe the clinically relevant outcome of interest (e.g., regaining consciousness).

Two linked challenges arise for machine learning. %
First, it complicates training. For ease of exposition, throughout the paper we stick to the setting of binary classification, where we aim to predict whether a patient will have a favorable vs an unfavorable outcome. A naive strategy that has historically been applied when counterfactual guesses are not available for the uncertain cases is to just assume that uncertain cases all correspond to having the same label as the ``unfavorable'' outcome of the certain case in Figure~\ref{fig:label-indeterminacy-overview}. However, using this naive strategy for defining target labels ends up yielding classifiers that largely just learn patterns in when physicians previously took a limiting action rather than a patient's true %
recovery potential~\citep{wilkinson2009}.
Second, and more fundamentally, it complicates evaluation. Past studies that used the naive strategy for defining target labels either ignore the difference between certain and uncertain cases in evaluation metrics, or they primarily consider evaluation metrics computed only on the certain cases.

Our paper addresses both of the challenges stated above. In particular, we make three key contributions.
First, we formalize predictive modeling under treatment-induced label indeterminacy when we have target labels that differ between certain and uncertain cases. 
Second, we introduce a split evaluation framework that evaluates certain cases using standard observed-label metrics (e.g., AUROC, Brier score) and uncertain cases using a notion of agreement or ``alignment'' with clinical expert guesses of counterfactual outcomes. We show that in settings with treatment-induced label indeterminacy, model selection should be based on an explicit tradeoff between performance on certain cases and alignment on uncertain cases, because standard observed-label metrics alone can mask clinically consequential differences on uncertain cases.
Third, we empirically characterize the resulting tradeoff across various tabular models and neural networks, and introduce a simple two-parameter neural objective that makes movement along this tradeoff explicit:
the weight placed on observed poor-outcome cases and the degree of alignment to expert guesses on counterfactual outcomes of uncertain cases.

We position this work primarily as a problem formulation and evaluation framework for predictive modeling under the treatment-induced label indeterminacy setting, rather than as a novel algorithmic contribution. %
Our work highlights the importance of characterizing the tradeoff between a model's performance on certain vs uncertain cases. Specifically for evaluation and model training on uncertain cases, having access to clinical experts' guesses of counterfactual outcomes 
is necessary for directly studying uncertain-case alignment, \textcolor{black}{but collecting such assessments requires substantial clinical effort. We therefore present the framework as a way to use carefully elicited assessments when they are available, rather than as a requirement that every clinical deployment routinely obtain counterfactual expert labels.}

\subsection*{Generalizable Insights about Machine Learning in the Context of Healthcare}

Although we study this setting in post-cardiac-arrest prognostication, the same structure (that of treatment-induced label indeterminacy) arises whenever a clinical decision systematically determines which outcomes are observed. %
For example, this structure appears in other clinical contexts such as in traumatic brain injury, dialysis initiation choices in end-stage renal disease, and triage decisions that determine whether a patient receives a diagnostic workup whose result would serve as the outcome label \citep{leblanc2018, voorend2022, mullainathan2022}.

Three lessons generalize beyond this case study.  
First, in these settings, we suspect that characterizing the tradeoff between model performance on certain and uncertain cases is important, and that only focusing model evaluation on the certain cases may lead to misleading conclusions or possibly poor predictions on the uncertain cases. %
Second, expert guesses of counterfactual outcomes can serve as target labels for the uncertain cases, albeit as imperfect supervision (since the guesses are not necessarily correct and different clinical experts can disagree on the same uncertain case).
Third, the tradeoff between certain-case performance and uncertain-case alignment is not a nuisance to be minimized but an explicit design choice that practitioners must navigate based on the clinical costs of each failure mode.

\section{Related Work}

This work lies at the intersection of several bodies of literature: post-cardiac-arrest prognostication, selective labels, the use of expert assessments when clinically meaningful outcomes are unavailable, and calibration in high-stakes decision-making tasks. %

\subsection{Post-Cardiac-Arrest Prognostication}

The work most closely related to ours is \citet{schoeffer2025}, who introduce the concept of \emph{label indeterminacy} in post-cardiac-arrest prognostication. Their key observation is that when patients die following WLST, their counterfactual recovery potential becomes permanently unknowable, and different plausible ways of handling such cases during training can yield materially different predictions. Our work shares this motivating concern but asks a complementary question: rather than focusing on how to construct training labels for these patients, we ask how the predicted probability of recovery should be \emph{evaluated} when the target outcome is no longer observable.

Machine learning for neurological outcome prediction after cardiac arrest has used diverse signals, including EEG, ECG, imaging, and clinical covariates, to predict discharge outcomes \citep{shen2023,niu2025,kawai2025,kasturi2026}. These studies show that early prognostication is increasingly feasible, but they are typically trained and evaluated against observed hospital outcomes, even though treatment decisions can alter the clinical trajectories those outcomes reflect. In resuscitation science, this concern has often been framed as a self-fulfilling prophecy problem: pessimistic prognostic judgments can increase the chance that recovery potential is never observed \citep{de2023,elmer2023,callaway2025,schoeffer2025}. Our work builds on this literature but focuses on the evaluation problem that remains once this observability failure is recognized.

\subsection{Selective Labels, Dependent Censoring, and Counterfactual Outcomes}

A useful lens for our setting is the selective labels problem \citep{lakkaraju2017,kleinberg2018,wei2021a}. In this problem, outcomes are observed only conditional on prior human decisions, so evaluating a model only on the labeled subset can yield biased performance estimates because the observed cases form a decision-dependent, non-random sample of the population \citep{lakkaraju2017}. Our setting has this same structure: treatment decisions determine which patients retain an observed recovery outcome. In this sense, post-cardiac-arrest prognostication can be viewed as a particularly high-stakes selective-label setting.

Classical approaches to informative or dependent censoring, including inverse-probability weighting and imputation-based methods, offer principled corrections under assumptions linking label observability to measured covariates and the censoring mechanism \citep{robins2000,zhu2012,binder2014}. Related work in cardiac arrest has used sensitivity analyses, matched comparisons, and competing-risk models to study bias induced by WLST \citep{shen2023,shen2025stepwise,lagebrant2025,callaway2025}. These approaches are important for clarifying the causal structure of the problem, but they do not by themselves provide ground-truth evaluation for patients whose recovery outcome is never observed. In particular, when label observability depends on the same prognostic judgments the model is meant to support, and may additionally depend on unmeasured bedside factors, the assumptions needed to recover counterfactual labels are difficult to justify.

Selective-label evaluation work is therefore directly relevant to our main claim that standard observed-label evaluation is not enough. \citet{lakkaraju2017} propose an alternative evaluation strategy that avoids imputing all missing outcomes by exploiting heterogeneity across human decision-makers, while \citet{wei2021a} formulates selective labels as a policy-learning problem. Our goal is different. We do not seek to compare human and algorithmic accept/reject policies or to identify a single corrected label set for all patients. Instead, we treat this setting as an evaluation problem with two qualitatively different subsets: certain cases, for which standard supervised metrics remain appropriate, and uncertain cases, for which evaluation must rely on imperfect but clinically meaningful expert guesses of the counterfactual rather than observed target outcome.

\subsection{Expert Assessments of Counterfactual Outcomes}

When clinically meaningful outcomes are permanently unobservable, experts' assessments of what the counterfactual outcome would be can provide clinically meaningful information, but they should not be interpreted as ground-truth outcomes \citep{bojke2021developing,soares2024recommendations}. 
This differs from the classical multi-annotator setting, where multiple annotations are typically treated as noisy observations of a latent true label to be inferred \citep{dawid1979,raykar2010learning}. In our setting, however, the relevant label for uncertain cases is itself counterfactual and may never be observed. Expert disagreement therefore need not reflect annotator noise alone; it may instead reflect irreducible uncertainty about a genuinely unobservable clinical outcome. The assessments used in our experiments were collected in prior clinical work \citep{elmer2025recovery}; the contribution of the present paper is not the creation of that dataset, but the evaluation framework and model analysis enabled by it.

In the same broader setting, \citet{schoeffer2025} show that different plausible ways of handling such indeterminate cases can yield materially different models despite similar performance on patients with observed outcomes.
We build on that insight, but our goal is different. Rather than selecting a single pseudo-ground-truth label for uncertain cases, we convert expert assessments into probabilistic pseudo-labels and use them in a limited and explicit way: as expert-informed pseudo-labels for alignment on uncertain cases during training, and as expert-derived pseudo-labels for assessing model behavior on that subset during evaluation.

\subsection{Calibration and Subgroup-Aware Evaluation in Clinical Decision Support}

For clinical models that output probabilities, calibration is as important as discrimination because predicted risks are used near decision thresholds rather than only for ranking patients \citep{steyerberg2010,vancalster2019,guo2017}. This is especially important in neurological prognostication after cardiac arrest, where small differences near a very low recovery threshold can influence discussions about ongoing life-sustaining therapies. Decision-analytic tools such as decision-curve analysis likewise emphasize that the quality of a model depends on how probabilities align with action thresholds, not only on AUROC \citep{vickers2006decision}. At the same time, work on hidden stratification in medical ML shows that aggregate performance can conceal clinically meaningful failure modes in subpopulations \citep{oakden2020hidden}. Our setting combines these concerns: probability placement matters, and the subgroup where it matters most is precisely the subgroup for which standard label-based calibration is undefined. This motivates our split evaluation strategy: conventional discrimination and calibration on certain cases with usable observed outcomes, and expert-alignment-based assessment on uncertain cases whose recovery labels are indeterminate.

\section{Methods}

Our goal is to estimate a patient's probability of favorable neurological recovery had life-sustaining therapies been continued. We study this problem in a \emph{selective-label} setting induced by treatment-induced label indeterminacy: for some patients, the observed hospital outcome can be used as a supervised label for recovery prediction, whereas for others it cannot because treatment-limitation decisions or competing terminal events may obscure latent recovery potential.

Our approach has three components. First, we construct expert-derived pseudo-labels for uncertain cases. Second, we train a simple neural network classifier with two controllable mechanisms: reweighting of observed poor-outcome cases and direct alignment to expert-derived uncertain-case targets. Third, we evaluate models through the tradeoff between conventional performance on certain cases and clinically plausible probability estimates on uncertain cases.

\subsection{Problem Setup}
\label{sec:problem_setup}

Let $x_i \in \mathcal{X}$ denote the clinical covariates for patient $i$, measured within the prognostic time window, and let
\[
f_\theta(x_i) \in [0,1]
\]
denote the predicted probability of favorable neurological recovery.

A central challenge in this setting is that observed outcomes are not equally informative for all patients. In this retrospective cohort, we therefore partition patients into two disjoint subsets according to whether the observed hospital outcome is treated as a usable label for latent recovery potential.

The first subset, $\mathcal{C}$, contains \emph{certain} cases for which the observed hospital outcome is treated as a valid supervised label. For each $i \in \mathcal{C}$, we observe
$y_i \in \{0,1\} $,
where $y_i=1$ denotes favorable recovery and $y_i=0$ denotes unfavorable outcome. The exact operational outcome definition used in the analytic cohort is given in Section~\ref{sec:cohort} and Appendix~\ref{app:cohort_details}.

The second subset, $\mathcal{U}$, contains \emph{uncertain} cases. These are patients whose observed clinical course is not treated as a reliable label for latent recovery potential. In our setting, this corresponds to patients whose outcomes may have been shaped by withdrawal of life-sustaining therapies for perceived poor neurologic prognosis and/or by competing terminal events such as multisystem organ failure or rearrest. For these patients, observed non-recovery does not necessarily imply absence of recovery potential, so uncertain cases are excluded from standard supervised binary-outcome training.

This distinction is central to the problem. Training only on $\mathcal{C}$ does not merely reduce sample size; it anchors the scale for probability of recovery entirely on patients with cleanly observed labels. If uncertain cases differ systematically from certain cases, then standard evaluation on $\mathcal{C}$ may overstate how reliable the model is on the full clinical population.

\subsection{Expert Assessments for Uncertain Cases}
\label{sec:expert_assessment}

Each uncertain patient $j \in \mathcal{U}$ is associated with multiple expert assessments of recovery likelihood collected under a structured review protocol. Experts reviewed deidentified case summaries and answered a counterfactual question: if life-sustaining therapies had been continued, what would this patient's probability have been of regaining consciousness and surviving hospital discharge? 
\textcolor{black}{Experts selected one of seven ordered response categories, each defined by a verbal description and an associated probability interval, ranging from no chance of recovery to more likely than not to recover. The complete response scale is provided in Appendix~\ref{app:expert_mapping}.}

Let $s_{j,1}, \ldots, s_{j,m_j}$ denote the expert scores for uncertain patient $j$, where $m_j$ is the number of available assessments. Higher scores indicate greater expected recovery likelihood. We do not treat these assessments as ground-truth outcomes. Instead, we use them as expert-informed pseudo-labels about latent recovery potential in cases for which observed hospital outcomes are not considered reliable labels.

To place these assessments on the same scale as model outputs, we map each ordinal expert score to a probability reference value using a fixed mapping function $\phi(\cdot)$ and aggregate across raters:
\[
\tilde{p}_j
=
\frac{1}{m_j}
\sum_{k=1}^{m_j} \phi(s_{j,k}).
\]
The resulting $\widetilde p_j \in [0,1]$ is the patient-level expert-derived reference for uncertain patient $j$. It is used as a pseudo-label for uncertain-case alignment during training and as a reference value during uncertain-case evaluation. The expert response scale and the numerical mapping used to construct $\widetilde p_j$ are described in Appendix~\ref{app:expert_mapping}.

These pseudo-labels are used only for uncertain patients during model training and evaluation. We do not convert them into binary labels and then treat certain and uncertain cases as having the same type of target information (some baselines we consider later use such an approach). Instead, we view observed outcomes for certain cases and expert-derived references for uncertain cases as qualitatively different forms of supervision. The latter provide a target for aligning model outputs with aggregated expert belief when the relevant recovery outcome is not directly observable.

\subsection{Alignment-Based Classifier for Recovery}
\label{sec:align-method}

Our main model is a single-head neural predictor of favorable neurological recovery. The design goal is to preserve a single probability output across all experiments while allowing controlled adjustment of how strongly the model is influenced by (i) observed poor-outcome cases and (ii) expert-informed uncertain-case assessments.

\paragraph{Supervised loss on certain cases.}
The baseline neural net classifier is trained on certain cases only using weighted binary cross-entropy:
\[
\mathcal{L}_{\mathrm{certain}}(\theta)
=
\frac{1}{\sum_{i \in \mathcal{C}} w_i}
\sum_{i \in \mathcal{C}}
w_i \,
\mathrm{BCE}\!\left(f_\theta(x_i), y_i\right),
\]
where
\[
w_i =
\begin{cases}
\omega_{\mathrm{bad}}, & y_i = 0,\\
1, & y_i = 1.
\end{cases}
\]
Here $\omega_{\mathrm{bad}} \ge 1$ controls the relative emphasis placed on observed poor-outcome cases. When $\omega_{\mathrm{bad}}=1$, this reduces to the standard unweighted objective.

We include this weighting term because observed poor-outcome cases play a disproportionate role in shaping the low-probability region of the learned recovery scale, where uncertain cases are likely to lie. In this clinical setting, patients who undergo treatment limitation are often those judged to have very low recovery potential, even though that potential is not directly observable. Varying $\omega_{\mathrm{bad}}$ therefore provides a simple way to control how strongly the model is anchored toward observed poor outcomes, separate from any direct use of expert-informed uncertain-case targets.

\paragraph{Alignment loss on uncertain cases.}
To incorporate uncertain cases without assigning them binary labels, we introduce an auxiliary alignment term:
\[
\mathcal{L}_{\mathrm{align}}(\theta)
=
\frac{1}{|\mathcal{U}_E|}
\sum_{j \in \mathcal{U}_E}
\left(f_\theta(x_j) - \tilde{p}_j\right)^2,
\]
where $\mathcal{U}_E \subseteq \mathcal{U}$ denotes uncertain cases with expert assessments and $\tilde{p}_j$ is the aggregated pseudo-label defined above. This term encourages the predicted probability of recovery to align with expert assessment of recovery potential on uncertain patients, without treating those assessments as observed outcomes.

\paragraph{Full objective.}
The full training objective is
\[
\mathcal{L}(\theta)
=
\mathcal{L}_{\mathrm{certain}}(\theta)
+
\lambda_{\mathrm{align}} \mathcal{L}_{\mathrm{align}}(\theta),
\]
where $\lambda_{\mathrm{align}} \ge 0$ controls the strength of uncertain-case alignment. Setting $\lambda_{\mathrm{align}}=0$ recovers the standard observed-only neural model. Increasing $\lambda_{\mathrm{align}}$ shifts the learned probability scale toward greater agreement with expert-informed uncertain-case targets, while $\omega_{\mathrm{bad}}$ separately controls the influence of observed poor-outcome cases. Together, these two parameters provide a simple and interpretable way to trace a continuum from conventional supervised learning to stronger expert-informed alignment.

\subsection{Evaluation in the Selective-Label Setting}
\label{sec:tradeoff_eval}

Evaluation mirrors the selective-label structure of training. On certain cases, model performance is assessed against observed binary outcomes using standard discrimination and probability-accuracy metrics. On uncertain cases, model behavior is assessed against expert-derived pseudo-labels rather than binary labels. For each uncertain patient, we compare the predicted probability of recovery to the aggregated expert reference probability $\tilde{p}_j$ and summarize this discrepancy using mean absolute error.

This yields a two-axis evaluation framework. The first axis measures conventional predictive performance on \textbf{certain cases}, where observed outcomes are treated as valid labels. The second measures agreement with expert assessment of recovery potential on \textbf{uncertain cases}, where binary supervision is unavailable or inappropriate. %

Rather than collapsing these objectives into a single scalar metric, we study the tradeoff between them: accurate prediction on certain cases versus probabilities closer to the expert-derived reference values on uncertain cases. This framing reflects the structure of the problem. Uncertain patients can inform how model outputs are positioned relative to expert belief, but they should not be treated as if their latent recovery outcomes were directly observed.

Accordingly, the main research question is whether uncertain-case alignment improves model behavior in the uncertain cohort, and at what cost to standard performance in the certain cohort. Section~\ref{sec:experiments} defines the exact experimental protocol and metrics used to quantify this certain--uncertain tradeoff. Throughout, uncertain patients are never used as ground-truth recovery labels, and each uncertain patient is evaluated only through out-of-fold predictions from a model that did not train on that patient.

\section{Cohort and Expert-Assessment Data} \label{sec:cohort}
This section summarizes the analytic cohort, the certain--uncertain case partition used throughout the paper, and the structured clinical features used for modeling. Full cohort-construction details, descriptive statistics, and feature definitions are provided in Appendix~\ref{app:cohort_details} and Appendix~\ref{app:feature_details}. 

\paragraph{Analytic cohort.}
We evaluated the proposed approach on a retrospective analytic cohort derived from a single-center post-cardiac-arrest registry of adults who were unresponsive to verbal commands after resuscitation from cardiac arrest and hospitalized between January~1, 2010, and July~31, 2022. Starting from 3,719 registry patients with complete outcome ascertainment, we excluded patients who were awake on arrival, died after treatment withdrawal for non-neurological reasons, had traumatic or primary neurologic etiologies, or were younger than 18~years, yielding a final analytic cohort of 2,497 patients. 

Using the certain and uncertain case definitions introduced in Section~\ref{sec:problem_setup}, the analytic cohort was partitioned into 1,068 certain cases and 1,429 uncertain cases.
Among certain cases, 726 had favorable observed neurological recovery and 342 had unfavorable observed outcomes. For certain cases, the supervised target was a binary indicator of neurological recovery at hospital discharge: $y=1$ for favorable recovery and $y=0$ for unfavorable outcome. Favorable recovery was defined as regaining consciousness and surviving to hospital discharge with CPC\footnote{The Cerebral Performance Category (CPC) scale is a five-point, Glasgow-Pittsburgh scoring system used to assess neurological outcome, primarily after cardiac arrest and resuscitation. It rates functional survival from CPC 1 (normal functioning) to CPC 5 (brain death)}~1--3. Although many cardiac-arrest studies treat CPC~3 as a poor outcome, we followed the prior peer-reviewed expert-assessment study underlying this dataset \citep{elmer2025recovery} in classifying it as favorable, because neurological function in these patients often continues to improve after discharge from acute care. Unfavorable outcomes comprised patients discharged with CPC~4 or those who progressed to death by neurologic criteria.

Uncertain cases were patients whose observed clinical course was not treated as a valid supervised label for latent recovery potential. They were excluded from supervised binary-outcome training and used only through expert-derived reference information, as described in Section~\ref{sec:expert_assessment}. All uncertain cases had available expert assessments collected under a previously peer-reviewed structured review protocol \citep{elmer2025recovery}, with a median of 3 ratings per patient (IQR~3--3; range 3--5). %

\paragraph{Feature set.}
Model inputs consisted of structured demographic and early clinical features routinely available during the prognostic time window. These included age, sex, transfer status, out-of-hospital arrest, cardiac catheterization, initial arrest rhythm, Pittsburgh Cardiac Arrest Category (PCAC), arrest etiology, and FOUR motor score.

\section{Experiments}  \label{sec:experiments}

Our experiments are designed to test a key question: how do different label-handling strategies behave under treatment-induced label indeterminacy? Because observed outcomes are valid only for the certain subset, evaluation must distinguish between conventional predictive performance on certain cases and alignment with expert-derived reference probabilities on uncertain cases. We therefore compare models along these two axes, using both aggregate tradeoff plots and subgroup-level prediction distributions to assess how alternative training strategies affect the scale for the estimated probability of recovery.

\subsection{Experimental Setup}
All experiments used 5-fold cross-validation with separate out-of-fold evaluation for the certain and uncertain cohorts. Certain cases were split with stratification on the observed recovery label, and standard supervised performance was assessed only on held-out certain cases. Uncertain cases were split independently. Any model that used uncertain cases during training was allowed to use only uncertain training folds and was evaluated on the held-out uncertain fold, so that every uncertain patient received exactly one out-of-fold prediction from a model that did not train on that patient.

We evaluated models along the two axes introduced in Section~\ref{sec:tradeoff_eval}. On certain cases, we compute AUROC and Brier score against the observed binary outcomes, with Brier score serving as the primary probability-sensitive metric. On uncertain cases, we compare predicted recovery probabilities with the patient-level mean of the mapped expert assessments using mean absolute error (MAE).
For the primary experiments, each ordinal expert response category is
assigned a prespecified probability reference value. This primary mapping uses the upper endpoint of each bounded response interval and assigns the open-ended highest category, ``more likely than not,'' a value of $0.75$.
The resulting patient-level expert mean is used both to train expert-informed models and to compute uncertain-case MAE. 
We additionally assess sensitivity to lower-valued and midpoint representations of the response categories, with the treatment of the open-ended highest category specified in Appendix~\ref{app:expert_mapping}.
Additional uncertain-case analyses, including subgroup summaries for expert groups A--C around the clinically important 1\% threshold, are reported in Appendix~\ref{app:full_results}.

We first compare representative tabular baselines. To anchor the comparison against prior work \citep{schoeffer2025} on a similar recovery-prediction problem with label indeterminacy, we include: an ``XGBoost observed-only'' model trained on certain cases only, and an ``XGBoost expert-average'' model that assigns each uncertain patient the mean mapped expert probability as a pseudo-training target. 
These two baselines represent, respectively, a conventional supervised strategy and a simple expert-informed alternative. The full tabular baseline suite, including additional XGBoost, random forest, and TabPFN \citep{grinsztajn2026tabpfn} variants, is reported in Appendix~\ref{app:baseline_details}.

Our primary model family is the neural recovery model introduced in Section~\ref{sec:align-method}. All neural models share the same architecture and training procedure and differ only in two method parameters: the poor-outcome weight $\omega_{\mathrm{bad}}$ and the uncertain-case alignment strength $\lambda_{\mathrm{align}}$. The setting $(\omega_{\mathrm{bad}}=1, \lambda_{\mathrm{align}}=0)$ corresponds to a standard observed-only neural model trained on certain cases without additional weighting or uncertain-case alignment. Increasing $\omega_{\mathrm{bad}}$ emphasizes observed poor-outcome cases, while increasing $\lambda_{\mathrm{align}}$ encourages closer agreement with expert-derived uncertain-case reference targets. Full implementation and tuning details are provided in Appendix~\ref{app:tuning_details}.

For visualization and subgroup analysis, we highlight five representative models spanning the main strategies studied in this paper: (i) XGBoost observed-only, (ii) XGBoost expert-average, (iii) a neural observed-only model with $(\omega_{\mathrm{bad}}=1, \lambda_{\mathrm{align}}=0)$, (iv) a neural moderately aligned model $(\omega_{\mathrm{bad}}=8, \lambda_{\mathrm{align}}=0.5)$, and (v) a neural strongly aligned model $(\omega_{\mathrm{bad}}=32, \lambda_{\mathrm{align}}=4)$. These representative models are used only for interpretation and visualization; the main quantitative conclusions are based on the full neural sweep together with the tabular baselines.

\subsection{Aggregate certain--uncertain tradeoff}
\label{sec:results_tradeoff}

We first characterize the aggregate tradeoff between standard performance on certain cases and expert-aligned behavior on uncertain cases. The key empirical question is not which model maximizes a single scalar metric, but how different label-handling strategies move along the resulting certain--uncertain tradeoff frontier.

\begin{figure}[t]
    \centering
    \includegraphics[width=0.49\textwidth]{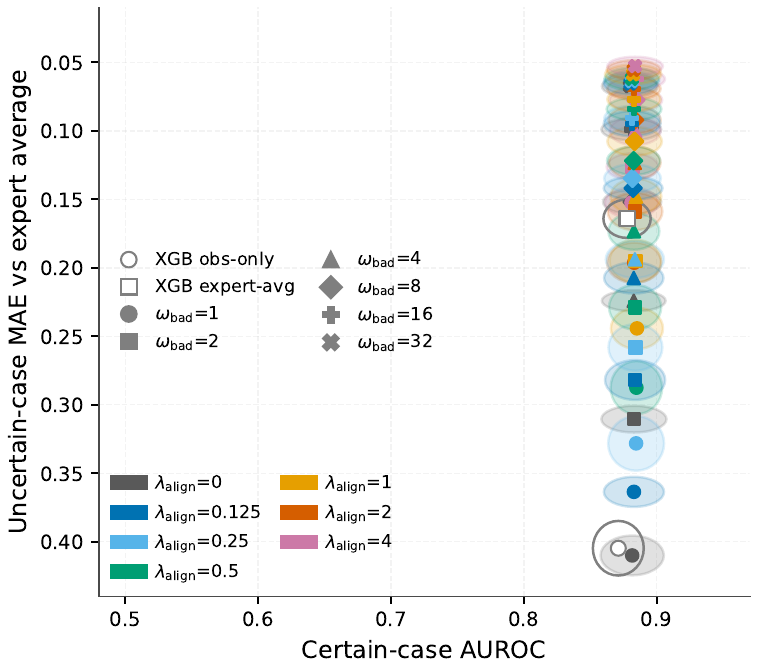}
    \hfill
    \includegraphics[width=0.49\textwidth]{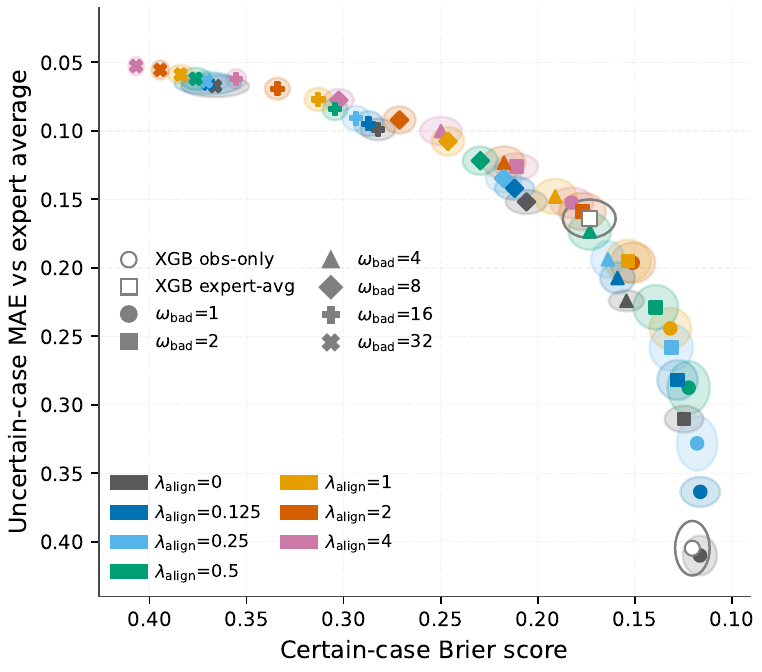}
    \caption{\textbf{Certain--uncertain tradeoff across model configurations.}
    Each point is one model configuration, with the point center showing the mean across the five outer folds and the ellipse radii indicating one standard deviation in each metric. Left: certain-case AUROC versus uncertain-case mean absolute error (MAE) relative to the expert-average reference probability. Right: certain-case Brier score versus uncertain-case MAE. Tabular baselines are shown for comparison; neural models vary the poor-outcome weight $\omega_{\mathrm{bad}}$ and uncertain-case alignment strength $\lambda_{\mathrm{align}}$. Similar AUROC values can correspond to widely different uncertain-case behavior, whereas the Brier--MAE view reveals a clear frontier between certain-case probability accuracy and uncertain-case expert alignment.}
    \label{fig:trade-off-scatter-main-paper}
\end{figure}

Figure~\ref{fig:trade-off-scatter-main-paper} shows that certain-case performance alone does not determine model behavior on uncertain patients. Across model configurations, certain-case AUROC changes relatively little, whereas uncertain-case MAE spans a wide range. At the same time, the right panel shows that lower uncertain-case MAE is generally associated with worse certain-case Brier score. In other words, models with similar discrimination on the observed-label cohort can place uncertain patients very differently on the scale for probability of recovery, and those changes are accompanied by a clear tradeoff in certain-case probability accuracy. This reliability gap is largely invisible if evaluation is restricted to AUROC or other ranking-based metrics alone.

The representative XGBoost baselines already exhibit this pattern. The observed-only XGBoost model achieves strong certain-case performance but poor uncertain-case alignment, whereas the expert-average XGBoost model substantially reduces uncertain-case error by sacrificing certain-case probability accuracy. This suggests that the tradeoff is induced by the label structure itself rather than by the neural architecture alone.

The neural model family makes this tradeoff especially transparent. The observed-only neural anchor, trained with $(\omega_{\mathrm{bad}}=1, \lambda_{\mathrm{align}}=0)$, achieves strong certain-case performance but assigns substantially higher probabilities to uncertain cases than models that incorporate expert information. Increasing the poor-outcome weight $\omega_{\mathrm{bad}}$ moves uncertain-case predictions downward and improves uncertain-case MAE, but does so at a monotone cost in certain-case Brier score while AUROC changes comparatively little. Increasing $\lambda_{\mathrm{align}}$ provides a more direct mechanism for pulling predictions toward the expert-informed uncertain-case reference probabilities, again at the cost of worse certain-case Brier score. Taken together, these results show that both $\omega_{\mathrm{bad}}$ and $\lambda_{\mathrm{align}}$ move models along the same frontier, but through different levers: poor-outcome weighting strengthens the influence of the observed low-probability region, whereas explicit alignment directly anchors uncertain-case predictions to aggregated expert belief. The full 42-model neural heatmaps in Appendix~\ref{app:full_results} show that this tradeoff is smooth across the joint $(\omega_{\mathrm{bad}}, \lambda_{\mathrm{align}})$ sweep rather than isolated to a few selected operating points.

\begin{table}[t]
    \centering
    \begin{adjustbox}{max width=0.95\textwidth}
    \begin{tabular}{lccc}
        \toprule
        Model & AUROC (certain) $\uparrow$ & Brier (certain) $\downarrow$ & MAE (uncertain) $\downarrow$ \\
        \midrule
        (i) XGBoost observed-only & 0.871 $\pm$ 0.019 & 0.120 $\pm$ 0.009 & 0.405 $\pm$ 0.020 \\
        (ii) XGBoost expert-average & 0.878 $\pm$ 0.018 & 0.173 $\pm$ 0.014 & 0.164 $\pm$ 0.014 \\
        (iii) Neural observed-only $(1, 0)$ & 0.882 $\pm$ 0.024 & 0.116 $\pm$ 0.009 & 0.410 $\pm$ 0.014 \\
        (iv) Neural moderate alignment $(8, 0.5)$ & 0.883 $\pm$ 0.020 & 0.230 $\pm$ 0.009 & 0.122 $\pm$ 0.010 \\
        (v) Neural strong alignment $(32, 4)$  & 0.884 $\pm$ 0.021 & 0.407 $\pm$ 0.004 & 0.053 $\pm$ 0.007 \\
        \bottomrule
    \end{tabular}
    \end{adjustbox}    
    \caption{\textbf{Representative models illustrating the certain--uncertain tradeoff.}
    Reported values are mean $\pm$ standard deviation across the five outer folds. Lower Brier score indicates better probability accuracy on certain cases, whereas lower uncertain-case MAE indicates closer agreement with expert-informed reference probabilities on uncertain cases. The table illustrates the tradeoff visible in Figure~\ref{fig:trade-off-scatter-main-paper} rather than identifying a universally optimal model. For the neural models, parentheses report $(\omega_{\mathrm{bad}}, \lambda_{\mathrm{align}})$.}
    \label{tab:representative_tradeoff}
\end{table}

Table~\ref{tab:representative_tradeoff} makes the scale of this tradeoff concrete. For example, the standard observed-only neural model and the strongly aligned neural model have nearly identical certain-case AUROC (0.882 vs.\ 0.884), yet their certain-case Brier scores and uncertain-case MAE differ dramatically (0.116 vs.\ 0.407 and 0.410 vs.\ 0.053, respectively). Thus, ranking performance (like AUROC) alone would suggest little difference between these models, even though they behave very differently on both the certain-case probability scale and the uncertain cohort. Full tradeoff tables across all evaluated runs, together with the broader tabular baseline family, are reported in Appendix~\ref{app:full_results}.

\subsection{Prediction-scale behavior on uncertain patients}
\label{sec:results_distributions}

We next examine how the aggregate tradeoff in Figure~\ref{fig:trade-off-scatter-main-paper} manifests in the predicted probability distributions of representative models. This analysis clarifies not only whether models improve uncertain-case agreement on average, but also how they reshape the scale for probability of recovery across clinically distinct subgroups.

\begin{figure}[t]
    \centering
      \includegraphics[width=1\linewidth]{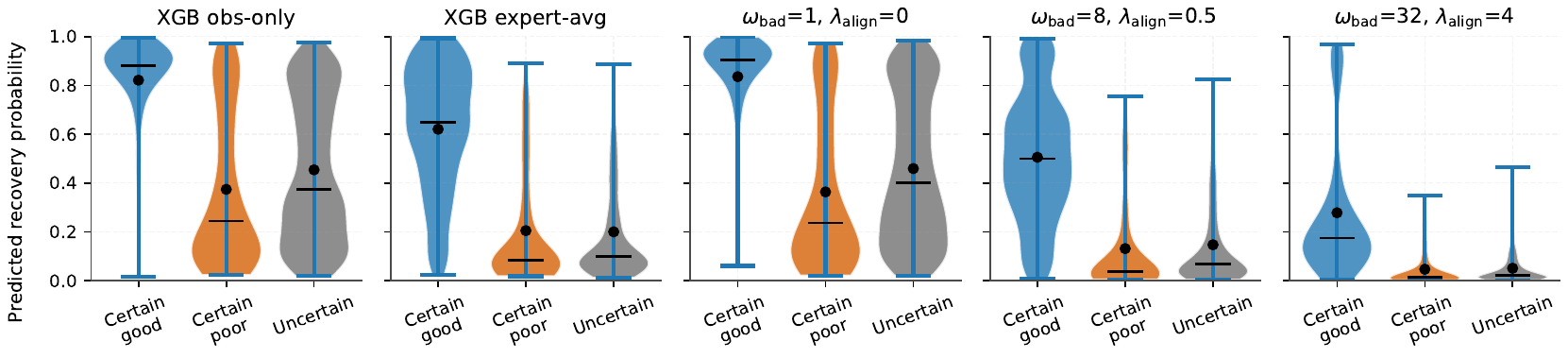}
      \vspace{1.em}
      \includegraphics[width=1\linewidth]{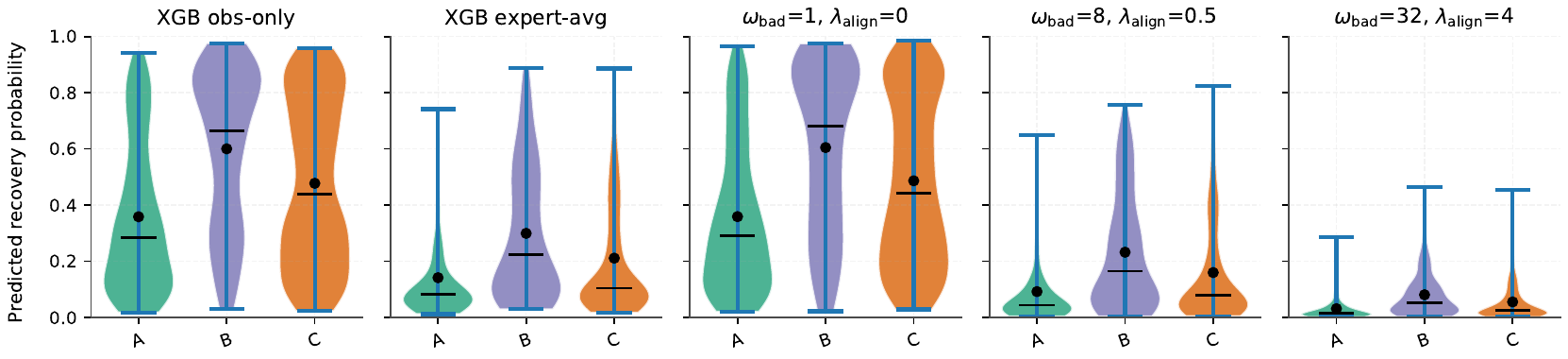}
    \vspace{-2.em}
    \caption{\textbf{Predicted recovery probabilities for representative models.}
    Violin plots show out-of-fold predicted probabilities; black dots and
    horizontal bars denote subgroup means and medians. Top: certain good,
    certain poor, and uncertain cases. Bottom: uncertain cases divided
    into group A (all expert assessments $\leq 1\%$), group B (all expert
    assessments $>1\%$), and group C (experts disagree across the $1\%$
    threshold). 
    Moderate alignment shifts uncertain cases toward the
    expert-assessed low-probability region while preserving greater subgroup
    separation than the strongly aligned model.}
    \label{fig:pred-distribution-main-paper}
\end{figure}

Figure~\ref{fig:pred-distribution-main-paper} illustrates how these aggregate differences appear on the prediction scale. In the observed-only models, uncertain cases receive relatively high recovery probabilities and overlap substantially with certain good cases. This suggests that standard supervised learning does not place uncertain patients in a region more consistent with the aggregate expert assessments: they are not clearly separated from certain good cases, yet neither are they anchored to the lower-probability region associated with poor expected recovery. Incorporating expert information shifts uncertain-case predictions downward. The moderately aligned neural model yields an intermediate regime in which uncertain cases are more clearly separated from certain good cases without collapsing all predictions toward zero. By contrast, the strongly aligned neural model compresses uncertain cases and certain poor cases into a narrow low-probability region and also lowers predictions for certain good cases, consistent with the deterioration in certain-case Brier score seen in Figure~\ref{fig:trade-off-scatter-main-paper} and Table~\ref{tab:representative_tradeoff}.

The subgroup plots further show how alignment affects the three expert-defined groups introduced in Figure~\ref{fig:pred-distribution-main-paper}.
In the observed-only models, all three groups receive relatively elevated and weakly separated predictions. With moderate alignment, predictions for group A shift closer to the near-zero region while group B remains somewhat higher, and group C occupies an intermediate band. Under aggressive alignment, however, all three groups are compressed downward, including group B. This suggests that stronger alignment improves average uncertain-case agreement partly by collapsing clinically heterogeneous uncertain patients onto a uniformly low-probability scale. Additional groupwise threshold-relative summaries are reported in Appendix~\ref{app:groupwise_uncertain_results}.

Taken together, these results support three conclusions. First, strong performance on certain cases does not guarantee sensible behavior on uncertain cases. Second, AUROC alone is not sufficient to characterize model quality in this setting: the main failure mode of observed-only learning is probability misplacement, which becomes visible only when probability-sensitive metrics such as Brier score and uncertain-case alignment are also examined. Third, expert-guided uncertain-case alignment provides a practical mechanism for improving predictions in the indeterminate cohort, but only by moving along an explicit tradeoff frontier rather than by dominating standard supervised learning on all axes. Additional distribution plots for the broader baseline family and full uncertainty-focused subgroup analyses are reported in Appendix~\ref{app:full_results}.

\section{Discussion}

This study shows that treatment-induced label indeterminacy creates a hidden evaluation gap in clinical prediction. In post-cardiac-arrest prognostication, the patients for whom prognostic support is most consequential are also those for whom the clinically meaningful recovery label is no longer directly observable. Our results show that standard supervised evaluation is therefore incomplete: models with similar performance on certain cases can assign substantially different recovery probabilities to uncertain patients.

A central implication is that the main failure mode in this setting is miscalibration of classifier probabilities. In particular, across the models evaluated, AUROC changed comparatively little on the certain cases, whereas certain-case Brier score and uncertain-case agreement varied substantially. For a clinical task in which decisions may hinge on very low favorable outcome probability thresholds, this distinction matters. The key object is therefore not a single best model, but the certain--uncertain tradeoff frontier. Better agreement with expert-informed uncertain-case assessments generally came at the cost of worse probability accuracy on certain cases, making model selection an explicit design choice rather than a hidden consequence of standard supervised learning.

\paragraph{Extensions.} The analysis here treats each patient as a single static snapshot and produces one recovery probability per patient. A natural extension is to model the evolving clinical trajectory, producing predictions that update as new observations arrive during the ICU stay. A complementary direction is to broaden the feature set: we deliberately restricted the main experiments to a compact set of covariates that would be collected early on for a patient so that the certain–uncertain tradeoff would not be confounded by feature-engineering choices, but preliminary experiments suggest that incorporating vasopressor exposure improves performance, and continuous EEG is another natural candidate given its established role in post-cardiac-arrest prognostication. 

\paragraph{Limitations.} The analysis is retrospective and based on a single-center cohort, so the certain/uncertain partition and the resulting tradeoff may partly reflect local practice patterns. The uncertain-case target labels come from expert guesses of counterfactual outcomes, which might not actually be accurate, so better alignment with these target labels should not be interpreted as recovery of ground-truth counterfactual probabilities. More broadly, the certain/uncertain partition is itself an operationalization of label usability in this cohort, not a universally fixed categorization that will transfer unchanged across institutions or settings. We therefore do not claim to identify true counterfactual recovery probabilities or to solve label indeterminacy in general.

\acks{This work was supported by NSF CAREER award \#2047981, and by grants R01NS124642 and UL1TR001857 from the National Institutes of Health. The authors thank the anonymous reviewers for helpful feedback.}
\newpage
\bibliography{our}

\newpage
\appendix

\section{Cohort Construction and Study Setting}
\label{app:cohort_details}

This appendix provides the study-setting details, cohort-construction pipeline, case definitions, and descriptive characteristics that support Section~\ref{sec:cohort} of the main paper.

\subsection{Study Setting}
This study used data from a prospective, single-center registry of adults
(aged $\geq$18~years) who were unresponsive to verbal commands after
resuscitation from cardiac arrest and hospitalized at a large medical center in the Northeastern US between January~1, 2010, and July~31, 2022.

\subsection{Cohort Construction}

The analytic cohort was derived from the institutional post-cardiac-arrest registry. Starting from 3,719 registry patients with complete outcome ascertainment, we applied the following inclusion and exclusion criteria, consistent with the prior peer-reviewed expert-assessment study underlying this dataset \citep{elmer2025recovery}.

\paragraph{Inclusion criteria.}
Adults (aged $\geq 18$~years) who were unresponsive to verbal commands after resuscitation from cardiac arrest and hospitalized at the study site.

\paragraph{Exclusion criteria.}
\begin{itemize}
    \item \textbf{Awake on arrival} ($n=726$): patients whose initial FOUR motor score indicated that they were following commands, i.e., they were not comatose on presentation.
    \item \textbf{Treatment withdrawal for non-neurological reasons} ($n=337$): patients who died after withdrawal of life-sustaining therapies based on non-neurological considerations (e.g., preexisting advance directives unrelated to neurological prognosis or a Do-Not-Resuscitate (DNR) order).
    \item \textbf{Incompatible etiologies} ($n=154$): patients whose cardiac arrest resulted from trauma or primary neurologic etiologies because prognostication in these patients differs
    substantially from the general post-arrest cohort.
    \item \textbf{Age $<$18~years} ($n=5$).
\end{itemize}

\begin{figure}[htbp]
    \centering
      \includegraphics[width=1\linewidth]{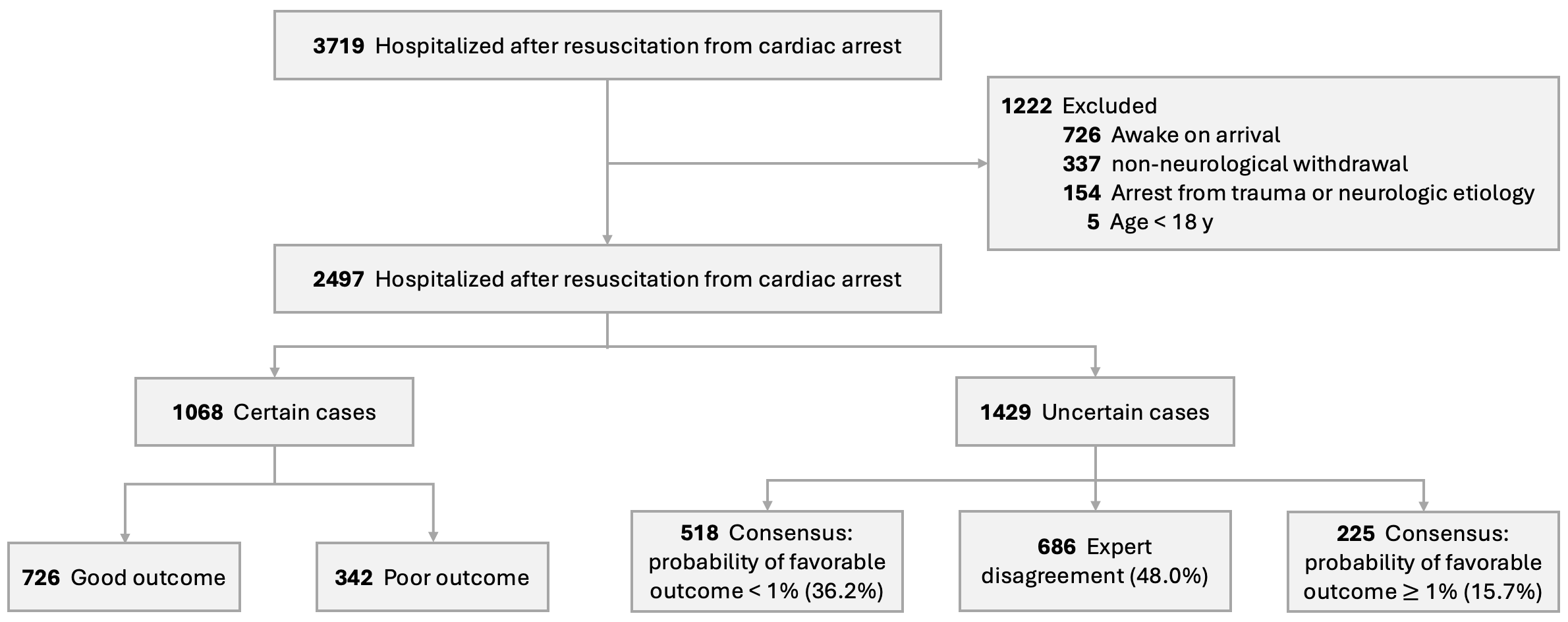}
    \vspace{-1.em}
    \caption{\textbf{Cohort-construction flow diagram.} Starting from 3,719 registry patients with complete outcome ascertainment, exclusions yielded a final analytic cohort of 2,497 patients, partitioned into 1,068 certain cases and 1,429 uncertain cases.}
    \label{fig:cohort-selection-flow}
\end{figure}

After applying all exclusion criteria, 2,497 patients remained in the analytic cohort. Figure~\ref{fig:cohort-selection-flow} summarizes the cohort-construction pipeline.

\subsection{Certain and Uncertain Case Definitions}

\textbf{Certain cases} ($n = 1068$) were those for whom the observed hospital outcome was treated as a usable supervised label for recovery prediction. This group included:
\begin{itemize}
    \item Good (or favorable) recovery ($y = 1$, $n = 726$): patients who regained consciousness and survived to hospital discharge with CPC~1 to~3.
    \item Poor (or unfavorable) recovery ($y = 0$, $n = 342$): patients discharged with CPC~4 or progressed to death by neurologic criteria.
\end{itemize}

\noindent \textbf{Uncertain cases} ($n = 1429$) were those for whom the observed clinical outcome was considered uncertain and thus not treated as a valid label for latent recovery potential.
Among the 1,429 uncertain cases, 867 (60.7\%) followed a WLST, while 562 (39.3\%) experienced multisystem organ failure or rearrest that prevented the continued-treatment neurological trajectory from being observed.

\noindent %
Uncertain cases were never treated as having observed binary recovery labels. Their inclusion in training depended on the label-handling strategy: some strategies excluded them, whereas others incorporated them using pseudo-labels or imputed targets.

\subsection{Descriptive characteristics}

Table~\ref{tab:patient-summary-stats} summarizes the demographic and clinical characteristics of the analytic cohort stratified by favorable observed recovery, unfavorable observed recovery, and uncertain recovery label.

\begin{table}[htbp]
  \centering
  \caption{\textbf{Patient characteristics summary statistics for the analytic cohort.} For binary covariates, values are reported as percentages within each cohort subgroup. For age, values are reported as means. The three outcome-oriented columns correspond to the modeling partition used in this paper: patients with favorable observed recovery, patients with unfavorable observed recovery, and patients with uncertain recovery labels.}
  \begin{adjustbox}{max width=0.99\textwidth}
  \begin{tabular}{lcccc}
    \toprule
    ~ & Total & Favorable & Unfavorable & Uncertain \\
    \midrule
    Number of subjects & 2497 & 726 & 342 & 1429 \\
    Percentage & 100\% & 29.1\% & 13.7\% & 57.2\% \\
    \midrule
    Age (yr) & 57.7 & 56.6 & 49.2 & 60.3 \\
    Female & 38.8\% & 34.0\% & 46.8\% & 39.4\% \\
    Out-of-hospital arrest & 82.0\% & 77.0\% & 92.7\% & 82.0\% \\
    Referral from outside facility & 62.1\% & 53.7\% & 68.4\% & 64.9\% \\
    Cardiac catheterization & 20.8\% & 41.0\% & 10.8\% & 12.9\% \\
    \midrule
    Initial rhythm & ~ & ~ & ~ & ~ \\
    \quad VF/VT   & 28.7\% & 49.4\% & 16.1\% & 21.1\% \\
    \quad PEA      & 34.6\% & 30.2\% & 31.9\% & 37.6\% \\
    \quad Asystole & 29.6\% & 12.5\% & 44.2\% & 34.7\% \\
    \quad Other    & 6.7\% & 6.6\% & 7.6\% & 6.6\% \\
    \quad Unknown               & 0.4\% & 1.2\% & 0.3\% & 0.0\% \\
    \midrule
    Pittsburgh Cardiac Arrest Category (PCAC) & ~ & ~ & ~ & ~ \\
    
    \quad Level I - Awake & 4.3\% & 14.9\% & 0.0\% & 0.0\% \\
    \quad Level II - Coma with Mild Dysfunction & 21.8\% & 46.1\% & 14.0\% & 11.3\% \\
    \quad Level III - Coma with Moderate/Severe Dysfunction & 10.6\% & 14.9\% & 5.6\% & 9.6\% \\
    \quad Level IV - Coma with Brainstem Reflex Loss & 50.0\% & 8.7\% & 69.0\% & 66.5\% \\
    \quad Unknown    & 13.3\% & 15.4\% & 11.4\% & 12.7\% \\
    \midrule
    Etiology & ~ & ~ & ~ & ~ \\
    \quad Cardiac & 29.2\% & 45.6\% & 16.1\% & 24.1\% \\
    \quad Metabolic or toxic & 17.1\% & 14.3\% & 35.4\% & 14.2\% \\
    \quad Respiratory & 21.7\% & 16.1\% & 26.6\% & 23.3\% \\
    \quad Shock or circulatory & 5.1\% & 3.4\% & 1.5\% & 6.9\% \\
    \quad Other & 26.8\% & 20.5\% & 20.5\% & 31.6\% \\
    \midrule
    FOUR motor score & ~ & ~ & ~ & ~ \\
    \quad Localizing to pain & 8.5\% & 22.3\% & 3.2\% & 2.7\% \\
    \quad Flexion response to pain & 15.8\% & 30.9\% & 10.2\% & 9.5\% \\
    \quad Extension response to pain & 2.2\% & 3.0\% & 1.5\% & 1.9\% \\
    \quad No response to pain & 49.1\% & 27.5\% & 70.8\% & 54.9\% \\
    \quad Myoclonus & 15.6\% & 0.7\% & 10.5\% & 24.4\% \\
    \quad Unable to determine & 8.9\% & 15.6\% & 3.8\% & 6.7\% \\
    \bottomrule
  \end{tabular}
  \end{adjustbox}
  \label{tab:patient-summary-stats}
\end{table}

\newpage
\section{Feature Definitions and Expert Assessments}
\label{app:feature_details}

This appendix documents the features, preprocessing, expert-assessment protocol, and expert-group definitions used in the main paper.

\subsection{Feature Set}

The final analytic table contained 20 model input columns after one-hot encoding of categorical predictors, with one category omitted from each encoded variable as the reference level. For readability, Table~\ref{tab:feature_dictionary} reports predictors at the level of clinical variables rather than expanded dummy columns.
\begin{table}[htbp]
\centering
\small
\caption{\textbf{Feature dictionary for predictors used in the analytic cohort.}
Categorical variables were one-hot encoded before modeling. %
}
\label{tab:feature_dictionary}
    \begin{adjustbox}{max width=1.0\textwidth}
    \begin{tabular}{p{0.24\linewidth} p{0.11\linewidth} p{0.32\linewidth} p{0.35\linewidth}}
    \toprule
    \textbf{Variable} & \textbf{Type} & \textbf{Coding} & \textbf{Clinical meaning} \\
    \midrule
    \multicolumn{4}{l}{\emph{Demographics and transfer status}} \\
    Age & Continuous & Years & Patient age at admission \\
    Sex & Binary & female/male indicator & Patient sex recorded in the clinical record \\
    Transfer status & Binary & transferred from outside facility: yes/no & Whether the patient was transferred from another hospital \\
    \midrule
    \multicolumn{4}{l}{\emph{Arrest circumstances and acute care}} \\
    Out-of-hospital arrest & Binary & yes/no & Whether the cardiac arrest occurred outside the hospital \\
    Cardiac catheterization & Binary & yes/no & Whether catheterization was performed during the acute hospitalization/prognostic window \\
    Initial rhythm & Categorical & VF/VT, PEA, asystole, other, unknown & Initial cardiac rhythm at arrest \\
    Pittsburgh Cardiac Arrest Category (PCAC) & Categorical & Level I, II, III, IV, unknown & Early post-arrest illness severity category \\
    Etiology & Categorical & cardiac, metabolic/toxic, respiratory, shock/circulatory, other & Suspected arrest etiology \\
    \midrule
    \multicolumn{4}{l}{\emph{Neurological examination}} \\
    FOUR motor score & Categorical & No response to pain/Extension response to pain/Myoclonus, Flexion response to pain, Localizing to pain, Unable to determine & Early motor response on the FOUR neurological examination \\
    \bottomrule
    \end{tabular}
    \end{adjustbox}
\end{table}

\subsection{Expert Assessments}
Each uncertain case had multiple expert assessments of recovery likelihood collected under a previously peer-reviewed structured review protocol \citep{elmer2025recovery}. Experts reviewed deidentified case summaries and estimated the probability that the patient would have awakened and survived to hospital discharge with CPC~1--3 had life-sustaining therapies been continued. The number of expert ratings per uncertain patient had a median of 3 (IQR~3--3; range 3--5). Expert assessments were never used as observed outcome labels.

\begin{table}[htbp]
\centering
\caption{\textbf{Expert response scale for uncertain-case recovery assessments.}}
\label{tab:expert_response_scale}
\begin{adjustbox}{max width=0.82\textwidth}
\begin{tabular}{p{0.15\textwidth} p{0.68\textwidth}}
\toprule
\textbf{Response category} & \textbf{Estimated probability of being awake and alive at hospital discharge (CPC~1--3)} \\
\midrule
1 & 0, no chance of awakening and survival to discharge \\
2 & $>0$ to 1\%, trivial chance of awakening and survival \\
3 & $>1$ to 5\%, very small chance of awakening and survival \\
4 & $>5$ to 10\%, small chance of awakening and survival \\
5 & $>10$ to 25\%, moderate chance of awakening and survival \\
6 & $>25$ to 50\%, good chance of awakening and survival \\
7 & $>50$\%, more likely than not to awaken and survive \\
\bottomrule
\end{tabular}
\end{adjustbox}
\end{table}

\begin{figure}[htbp]
    \centering
      \includegraphics[width=0.6\linewidth]{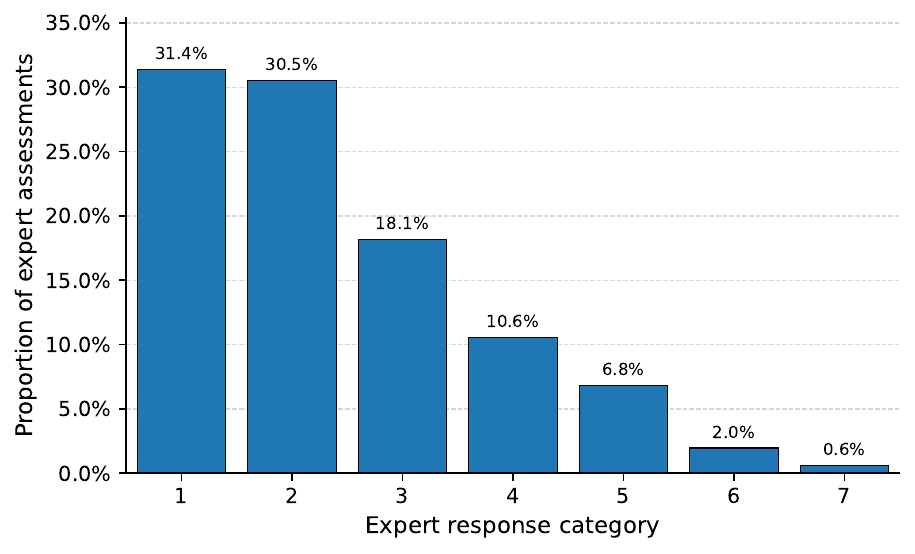}
    \caption{\textbf{Distribution of expert response categories in the uncertain cohort.}
    Bars show the proportion of all expert assessments assigned to each ordinal response category across uncertain cases. Ratings are concentrated in the lowest probability of recovery categories, with categories 1--2 accounting for the majority of assessments, while higher-probability responses are comparatively rare. This pattern indicates that most uncertain cases were judged to have low recovery potential, while still preserving meaningful heterogeneity within the uncertain subset.}
    \label{fig:expert-score-dist}
\end{figure}

Figure~\ref{fig:expert-score-dist} summarizes the empirical distribution of expert response categories in the uncertain cohort, showing that assessments are concentrated in the lowest probability of recovery categories but are not uniformly near zero.

\subsection{Expert score mapping}
\label{app:expert_mapping}

Experts selected one of the seven ordinal response categories shown in
Table~\ref{tab:expert_response_scale}. The experts did not provide point
probability estimates and were not asked to choose among the alternative
mappings described below. Rather, each response category corresponded to
a verbal description and an associated probability interval.

To use these responses as probabilistic reference targets, we assigned
each category a representative numerical value. The primary mapping used
for model training and the main evaluation assigns the upper endpoint of
each bounded interval. For the open-ended highest category, the elicitation specifies only that
the probability exceeds $0.50$. Our primary mapping assigns this category
a capped reference value of $0.75$. For sensitivity analysis, the
lower-endpoint value is $0.50$, and the midpoint value $0.625$ is defined
as the midpoint between $0.50$ and the primary cap of $0.75$; it is not
the midpoint of the full probability range $(0.50,1]$.
Table~\ref{tab:expert_score_mappings} gives the complete
numerical definitions.

\begin{table}[htbp]
    \centering
    \small
    \begin{adjustbox}{max width=1.0\textwidth}
    \begin{tabular}{ccccc}
        \toprule
        \textbf{Response}
        & \textbf{Probability interval}
        & \textbf{Lower endpoint}
        & \textbf{Interval midpoint}
        & \textbf{Primary mapping} \\
        \midrule
        1 & $0$             & 0.000 & 0.000 & 0.000 \\
        2 & $(0,0.01]$      & 0.000 & 0.005 & 0.010 \\
        3 & $(0.01,0.05]$   & 0.010 & 0.030 & 0.050 \\
        4 & $(0.05,0.10]$   & 0.050 & 0.075 & 0.100 \\
        5 & $(0.10,0.25]$   & 0.100 & 0.175 & 0.250 \\
        6 & $(0.25,0.50]$   & 0.250 & 0.375 & 0.500 \\
        7 & $>0.50$      & 0.500 & 0.625 & 0.750 \\
        \bottomrule
    \end{tabular}
    \end{adjustbox}
    \caption{\textbf{Numerical mappings of ordinal expert response
    categories.}
    The primary mapping is used for model training and the main
    uncertain-case evaluation. It uses the upper endpoint of each bounded
    interval and assigns the open-ended highest response category a value
    of $0.75$. The lower-endpoint and interval-midpoint mappings are used
    only for the fixed-prediction sensitivity analysis. These mappings
    are analytical representations applied after data collection and
    were not response options presented to the experts.}
    \label{tab:expert_score_mappings}
\end{table}

Let $\phi^{(\mathrm{P})}$, $\phi^{(\mathrm{L})}$, and
$\phi^{(\mathrm{M})}$ denote the primary, lower-endpoint, and
interval-midpoint mappings, respectively. For mapping
$r\in\{\mathrm{P},\mathrm{L},\mathrm{M}\}$, the patient-level
expert-derived reference is
\[
    \widetilde p_j^{(r)}
    =
    \frac{1}{m_j}
    \sum_{k=1}^{m_j}
    \phi^{(r)}(s_{j,k}).
\]
The primary experiments use $\widetilde p_j^{(\mathrm{P})}$. In the
sensitivity analysis, the fitted predictions from those experiments are
evaluated against $\widetilde p_j^{(\mathrm{L})}$ and
$\widetilde p_j^{(\mathrm{M})}$ without retraining.

\subsection{Expert groups around the 1\% threshold}
\label{sec:expert-group-abc}

For subgroup evaluation, uncertain cases were partitioned using the original ordinal response categories, independently of the numerical mapping used in the sensitivity analysis:
\begin{itemize}
    \item \textbf{Group A:} every expert selected category 1 or 2,
    corresponding to an elicited recovery probability at or below $1\%$;
    \item \textbf{Group B:} every expert selected category 3--7,
    corresponding to an elicited recovery probability above $1\%$;
    \item \textbf{Group C:} at least one expert selected category 1--2
    and at least one selected category 3--7.
\end{itemize}

The uncertain cohort contained 518 group-A cases, 225 group-B cases, and
686 group-C cases. Group membership was fixed from the original response
categories and was not recomputed under the alternative numerical
mappings. These groups were used only for evaluation and were never used
as observed recovery labels.

\section{Tabular Model Families and Label-Handling Strategies}
\label{app:baseline_details}

This appendix documents the broader baseline family used to contextualize the main-paper comparisons. In the main text, we report two representative XGBoost baselines to anchor the comparison against prior work. Here we describe the full tabular models label-handling suite used to verify that the main-paper certain--uncertain tradeoff is not specific to those two selected comparators.

\subsection{Label-handling strategies for tabular baselines}
We evaluated ten label-handling strategies adapted from prior work on label indeterminacy in post-cardiac-arrest recovery prediction. Table~\ref{tab:baseline_definitions} summarizes the correspondence between our implementation names and the original method family. Intuitively, these baselines span several distinct assumptions about how uncertain patients should contribute to training: ignoring them entirely, correcting only for selective observation, imputing pessimistic labels, borrowing labels from nearby certain patients, or incorporating expert-derived pseudo-labels under different aggregation rules.

All ten strategies were instantiated using random forest and XGBoost. TabPFN was evaluated under the eight strategies that do not require sample weighting. The two inverse-probability-weighted strategies were omitted for TabPFN because the TabPFN regressor interface used in our experiments does not natively accept a ``sample weight'' argument. For the random-forest implementation, we follow \cite{schoeffer2025} and use the random forest regressor, and the predicted values were clipped to $[0,1]$ to preserve a probability interpretation.

\begin{table}[htbp]
    \centering
    \small
    \begin{adjustbox}{max width=0.99\textwidth}
    \begin{tabular}{p{0.25\linewidth} p{0.56\linewidth} cc}
    \toprule
    \textbf{Model category} & \textbf{How uncertain cases enter training} & \textbf{RF/XGB}
        & \textbf{TabPFN} \\
    \midrule
    \texttt{observed only} &
    Train only on certain cases with observed binary recovery labels; uncertain cases are excluded. & $\checkmark$ & $\checkmark$\\

    \texttt{observed only ipw} &
    Train only on certain cases, but weight each certain case by the inverse estimated probability that its label would be observed. Uncertain cases are excluded from direct supervision. & $\checkmark$ & ---\\

    \texttt{assume uncertain zero} &
    Include all uncertain cases and assign each a pessimistic pseudo-label of $0$, effectively assuming that treatment-limited cases had no recovery potential. & $\checkmark$ & $\checkmark$\\

    \texttt{nearest neighbor} &
    For each uncertain case, impute the observed binary label of its nearest certain training case under the fold-local distance metric, and train on the union of certain and imputed uncertain rows. & $\checkmark$ & $\checkmark$\\

    \texttt{experts all} &
    For each uncertain case $j$ with $m_j$ expert-derived probabilities, duplicate the case once per expert probability and weight each duplicate by $1/m_j$. & $\checkmark$ & $\checkmark$\\

    \texttt{experts average} &
    For each uncertain case, assign a single pseudo-label equal to the mean of its mapped expert probabilities. & $\checkmark$ & $\checkmark$\\

    \texttt{experts max} &
    For each uncertain case, assign a single pseudo-label equal to the maximum of its mapped expert probabilities. & $\checkmark$ & $\checkmark$\\

    \texttt{experts agreement} &
    Include an uncertain case only if all mapped expert probabilities fall on the same side of the 1\% threshold; assign the case the maximum mapped expert probability within that agreeing set. Otherwise exclude it. & $\checkmark$ & $\checkmark$\\

    \texttt{experts agreement zero} &
    Include an uncertain case only if all mapped expert probabilities are at or below 1\%; assign the case the maximum mapped expert probability within that low-probability agreeing set. Otherwise exclude it. & $\checkmark$ & $\checkmark$\\

    \texttt{experts agreement zero ipw} &
    Construct uncertain-case labels as in \texttt{experts agreement zero}, then apply inverse-propensity weighting to the resulting training set to partially correct for selective label observability. & $\checkmark$ & ---\\
    \bottomrule
    \end{tabular}
    \end{adjustbox}
    \caption{\textbf{Label-handling strategies for tabular baselines and model-family coverage} The ten strategies are adapted from prior work \citep{schoeffer2025} on label indeterminacy in post-cardiac-arrest recovery prediction. The main paper highlights only two representative XGBoost baselines, while this appendix reports the broader comparator family.
    All ten strategies were evaluated using random forest (RF) and XGBoost (XGB). 
    TabPFN was evaluated under the eight strategies that do not require sample weighting. 
    The two inverse-probability-weighted (IPW) variants were omitted for TabPFN because the TabPFN regressor interface used in our experiments does not natively accept sample weights. The \texttt{experts all} strategy also relies on sample weights in its RF/XGB implementation ($1/m_j$ per duplicate, as defined above), but for TabPFN each uncertain case was instead duplicated using an unweighted subsample of at most 3 experts (randomly selected without replacement when more than 3 were available, affecting 528 of the uncertain patients), with all duplicates weighted equally at fit time; results for TabPFN \texttt{experts all} should therefore be interpreted as this approximation rather than the exact weighted-duplication strategy used for RF/XGB.
    }
    \label{tab:baseline_definitions}
\end{table}

These baselines are not intended to identify a single optimal tabular model. Rather, they provide a structured way to test whether the main-paper certain--uncertain tradeoff persists across a broader set of plausible supervision assumptions for uncertain patients.

\subsection{Model Families}

The label-handling strategies in Table~\ref{tab:baseline_definitions} were instantiated across three tabular model families. Random forest regression and XGBoost were evaluated under all ten strategies. The random-forest implementation most closely follows the model family used in \citet{schoeffer2025}, while XGBoost provides an additional gradient-boosted tree comparator under the same label-handling assumptions.

We additionally evaluated TabPFN as a modern non-tree tabular model. TabPFN was instantiated under the eight strategies that do not require sample weighting. The two inverse-probability-weighted variants were omitted because the TabPFN regressor interface used in our experiments does not natively accept sample weights. For strategies that incorporate expert-derived probabilities, TabPFN was fit through its regression interface so that both the binary observed labels for certain cases and the continuous pseudo-labels for uncertain cases could be used as training targets.

The main paper highlights the observed-only and expert-average configurations for XGBoost as representative tabular comparisons. Full results for all ten random-forest and XGBoost strategies and all eight TabPFN strategies are reported in Appendix~\ref{app:full_results}. The neural model studied in the main text constitutes a separate model family and uses the alignment-based objective introduced in Section~\ref{sec:align-method}.

\section{Training and Tuning Details}
\label{app:tuning_details}

This section provides the implementation details omitted from the main paper for readability. All preprocessing, tuning, model selection, and early stopping were confined to fold-local training data.

\subsection{Data Splitting}
All experiments used 5-fold outer cross-validation with a strict uncertain out-of-fold protocol. The certain cohort was split with stratification on the observed recovery label. The uncertain cohort was split independently into five folds. In fold $k$, any model that used uncertain cases during training was allowed to use only uncertain training folds and was evaluated on the held-out uncertain fold. Thus, every uncertain patient received exactly one out-of-fold prediction from a model that did not train on that patient.

Within each outer training fold, we created a shared inner validation split containing 15\% of the available training rows, stratified by certain versus uncertain source rows when feasible. This same inner split was shared across the neural model and the tuned tabular baselines to keep model-selection conditions aligned.

For the neural model, feature standardization was fit on the inner-training rows only and then applied to the corresponding validation and outer-fold test rows. Tabular models were trained on the prepared features without global rescaling. Any scaling used for auxiliary computations, such as distance calculations in nonparametric baselines, was also performed fold-locally.

\subsection{Tabular Model Hyperparameter Selection}

To ensure credible tabular baselines while keeping comparisons focused on label-handling assumptions, we tuned random forest, XGBoost, and TabPFN using inner-fold validation on certain cases only. Certain-case Brier score was the primary selection criterion, with AUROC used to break ties among configurations within a small absolute tolerance of the best Brier score. The selected hyperparameters for each model family were then reused across its label-handling variants.

The search spaces were:
\begin{itemize}
    \item \textbf{Random Forest:} \texttt{n\_estimators} $\in \{200, 500\}$, \texttt{max\_depth} $\in \{\texttt{None}, 8, 16\}$, \\ \texttt{min\_samples\_leaf} $\in \{1, 2, 5\}$;
    \item \textbf{XGBoost:} \texttt{eta} $\in \{0.03, 0.1\}$, \texttt{max\_depth} $\in \{3, 6\}$, \texttt{min\_child\_weight} $\in \{1, 5\}$, \\ \texttt{subsample} $\in \{0.8, 1.0\}$.
    \item \textbf{TabPFN:} \texttt{n\_estimators} $\in \{4,8,16\}$.
\end{itemize}

For TabPFN, we used \texttt{TabPFNRegressor} from the \texttt{tabpfn} Python package (version 8.0.6), corresponding to TabPFN v3~\citep{grinsztajn2026tabpfn}, with the default v3 regressor checkpoint. Ensemble size was the only tuned TabPFN hyperparameter; all other settings were held fixed across folds and label-handling strategies.

\subsection{Neural Training Configuration}

The single-head recovery MLP used two hidden layers of sizes 128 and 64, ReLU activations, dropout 0.2, Adam optimization, learning rate $10^{-4}$, batch size 64, weight decay $10^{-4}$, maximum 200 epochs, and early stopping patience 10.

The neural sweep varied the poor-outcome weight
\[
\omega_{\mathrm{bad}} \in \{1, 2, 4, 8, 16, 32\},
\]
and the uncertain-case alignment strength
\[
\lambda_{\mathrm{align}} \in \{0, 0.125, 0.25, 0.5, 1, 2, 4\}.
\]
This yielded 42 neural operating points spanning the continuum from conventional observed-only training to increasingly strong expert-guided alignment on uncertain cases.

\section{Extended Results}
\label{app:full_results}

This appendix provides the full empirical support for the main-paper
claims. We first examine the certain--uncertain tradeoff across the full
tabular and neural comparator set. We then assess the robustness and
interpretation of uncertain-case MAE through sensitivity to the numerical
mapping of the ordinal expert responses.
Finally, we report the full neural sweep and
additional prediction-distribution and expert-subgroup analyses.

\subsection{Full Tradeoff Across Model Families}
\begin{figure}[t]
    \centering
    \includegraphics[width=0.49\textwidth]{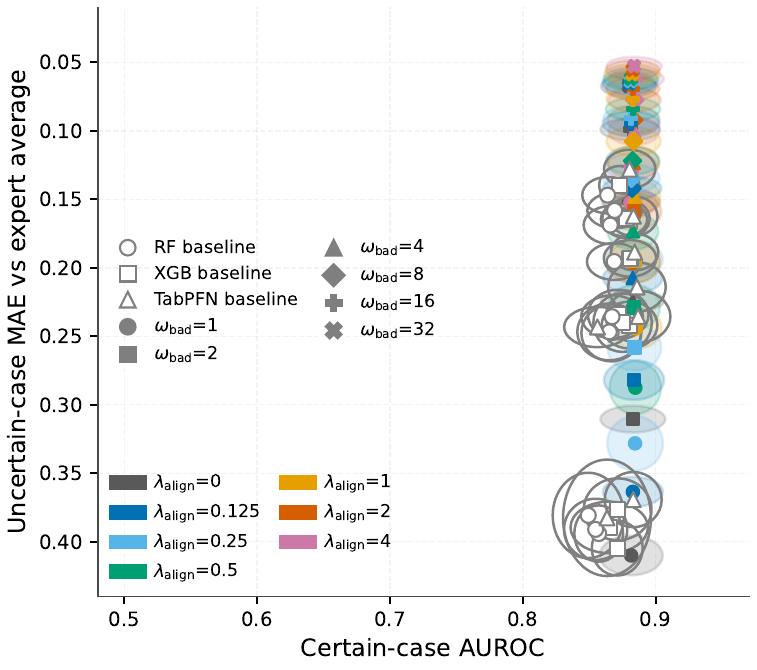}
    \hfill
    \includegraphics[width=0.49\textwidth]{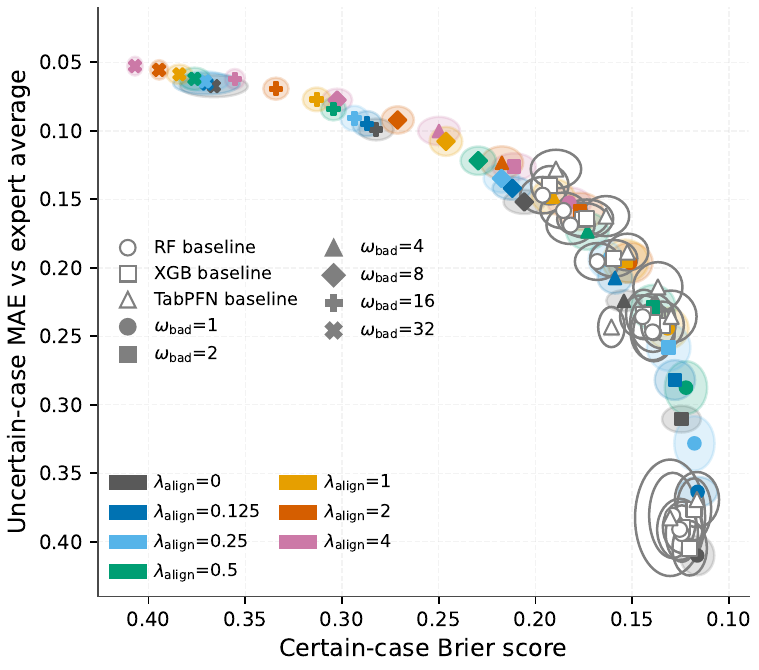}
    \caption{\textbf{Certain--uncertain tradeoff across the full set of model families.}
    This figure extends the aggregate tradeoff analysis in Figure~\ref{fig:trade-off-scatter-main-paper} by showing the full comparator set rather than the representative baselines highlighted in the main paper.
    Each point represents one model configuration, with point centers showing the mean across the five outer folds and ellipse radii showing one standard deviation in each metric. Left: certain-case AUROC versus uncertain-case mean absolute error (MAE) relative to the aggregated expert reference probability. Right: certain-case Brier score versus uncertain-case MAE. The broader tabular baseline family follows the same qualitative pattern as the representative models in the main paper: similar AUROC values can coexist with substantially different uncertain-case behavior, while improved uncertain-case alignment typically comes at the cost of worse certain-case Brier score.}
    \label{fig:trade-off-scatter-appendix}
\end{figure}

Figure~\ref{fig:trade-off-scatter-appendix} extends the main-paper tradeoff plots to the full comparator set. The same qualitative pattern persists: models that improve uncertain-case agreement generally do so by sacrificing certain-case probability accuracy, while AUROC changes comparatively little. This broader view also shows that the two XGBoost baselines highlighted in the main paper are representative of a wider family of tabular label-handling strategies rather than isolated examples.

Table~\ref{tab:full_tree_results} and \ref{tab:tabpfn_results} report the full numerical results for the tabular baseline family. Consistent with the main-paper tradeoff analysis, the observed-only baselines achieve relatively strong certain-case Brier scores but poor uncertain-case MAE, whereas expert-informed baselines improve uncertain-case agreement at the expense of certain-case calibration. No tabular baseline label-handling strategy dominates across both axes.

\begin{table}[htbp]
    \centering
    \begin{adjustbox}{max width=0.98\textwidth}
    \begin{tabular}{lccc}
    \toprule
    Model & AUROC (certain) & Brier (certain) & MAE (uncertain) \\
    \midrule
    XGB observed-only & 0.871 $\pm$ 0.019 & 0.120 $\pm$ 0.009 & 0.405 $\pm$ 0.020 \\
    XGB observed-only + IPW & 0.865 $\pm$ 0.029 & 0.124 $\pm$ 0.010 & 0.390 $\pm$ 0.019 \\
    XGB assume uncertain zero & 0.873 $\pm$ 0.015 & 0.193 $\pm$ 0.010 & 0.140 $\pm$ 0.014 \\
    XGB nearest neighbor & 0.871 $\pm$ 0.025 & 0.118 $\pm$ 0.009 & 0.376 $\pm$ 0.023 \\
    XGB experts all & 0.876 $\pm$ 0.017 & 0.176 $\pm$ 0.013 & 0.165 $\pm$ 0.012 \\
    XGB experts average & 0.878 $\pm$ 0.018 & 0.173 $\pm$ 0.014 & 0.164 $\pm$ 0.014 \\
    XGB experts max & 0.881 $\pm$ 0.020 & 0.160 $\pm$ 0.012 & 0.193 $\pm$ 0.013 \\
    XGB experts agreement & 0.872 $\pm$ 0.023 & 0.144 $\pm$ 0.009 & 0.234 $\pm$ 0.018 \\
    XGB experts agreement zero & 0.880 $\pm$ 0.021 & 0.134 $\pm$ 0.011 & 0.242 $\pm$ 0.017 \\
    XGB experts agreement zero + IPW & 0.875 $\pm$ 0.022 & 0.139 $\pm$ 0.012 & 0.240 $\pm$ 0.021 \\
    \midrule
    RF observed-only & 0.857 $\pm$ 0.018 & 0.124 $\pm$ 0.008 & 0.394 $\pm$ 0.021 \\
    RF Observed-only + IPW & 0.855 $\pm$ 0.024 & 0.125 $\pm$ 0.012 & 0.391 $\pm$ 0.022 \\
    RF assume uncertain zero & 0.864 $\pm$ 0.019 & 0.196 $\pm$ 0.012 & 0.147 $\pm$ 0.013 \\
    RF nearest neighbor & 0.849 $\pm$ 0.027 & 0.129 $\pm$ 0.012 & 0.381 $\pm$ 0.031 \\
    RF experts all & 0.869 $\pm$ 0.021 & 0.186 $\pm$ 0.012 & 0.158 $\pm$ 0.013 \\
    RF experts average & 0.866 $\pm$ 0.020 & 0.182 $\pm$ 0.012 & 0.169 $\pm$ 0.014 \\
    RF experts max & 0.869 $\pm$ 0.021 & 0.168 $\pm$ 0.012 & 0.195 $\pm$ 0.013 \\
    RF experts agreement & 0.867 $\pm$ 0.023 & 0.145 $\pm$ 0.012 & 0.236 $\pm$ 0.016 \\
    RF experts agreement zero & 0.867 $\pm$ 0.022 & 0.139 $\pm$ 0.011 & 0.248 $\pm$ 0.021 \\
    RF experts agreement zero + IPW & 0.865 $\pm$ 0.024 & 0.139 $\pm$ 0.011 & 0.247 $\pm$ 0.021 \\
    \bottomrule
    \end{tabular}
    \end{adjustbox}
    \caption{\textbf{Full random-forest and XGBoost baseline results.}
    This table provides the full numerical results corresponding to the tree-based points summarized in Figure~\ref{fig:trade-off-scatter-appendix}, and expands on the representative XGBoost baselines reported in Table~\ref{tab:representative_tradeoff} of the main paper. 
    Reported values are mean $\pm$ standard deviation across the five outer folds. Lower Brier score indicates better certain-case probability accuracy, whereas lower uncertain-case MAE indicates closer agreement with expert-informed reference probabilities on uncertain cases. Across both XGBoost and random forest, baselines that incorporate uncertain patients more aggressively tend to improve uncertain-case MAE while worsening certain-case Brier score.}
    \label{tab:full_tree_results}
\end{table}

\begin{table}[htbp]
    \centering
    \begin{adjustbox}{max width=0.98\textwidth}
    \begin{tabular}{lccc}
    \toprule
    Model & AUROC (certain) & Brier (certain) & MAE (uncertain) \\
    \midrule
    TabPFN observed-only & 0.883 $\pm$ 0.021 & 0.117 $\pm$ 0.011 & 0.369 $\pm$ 0.020 \\
    TabPFN assume uncertain zero & 0.880 $\pm$ 0.019 & 0.189 $\pm$ 0.013 & 0.128 $\pm$ 0.014 \\
    TabPFN nearest neighbor & 0.863 $\pm$ 0.033 & 0.130 $\pm$ 0.018 & 0.383 $\pm$ 0.043 \\
    TabPFN experts all & 0.856 $\pm$ 0.025 & 0.161 $\pm$ 0.007 & 0.243 $\pm$ 0.015 \\
    TabPFN experts average & 0.883 $\pm$ 0.018 & 0.164 $\pm$ 0.012 & 0.162 $\pm$ 0.015 \\
    TabPFN experts max & 0.884 $\pm$ 0.018 & 0.152 $\pm$ 0.011 & 0.189 $\pm$ 0.013 \\
    TabPFN experts agreement & 0.886 $\pm$ 0.022 & 0.137 $\pm$ 0.013 & 0.214 $\pm$ 0.018 \\
    TabPFN experts agreement zero & 0.887 $\pm$ 0.024 & 0.130 $\pm$ 0.013 & 0.236 $\pm$ 0.019 \\
    \bottomrule
    \end{tabular}
    \end{adjustbox}
    \caption{\textbf{TabPFN baseline results.} Reported values are mean $\pm$ standard deviation across the five outer folds. Lower Brier score indicates better certain-case probability accuracy, whereas lower uncertain-case MAE indicates closer agreement with expert-informed reference probabilities on uncertain cases.
    }
    \label{tab:tabpfn_results}
\end{table}

\subsection{Sensitivity to the Expert-Score Mapping}
\label{app:mapping_sensitivity}

The primary uncertain-case evaluation uses the numerical mapping defined
in Appendix~\ref{app:expert_mapping}. Because the original expert
responses were ordinal categories associated with probability intervals,
we assessed whether the comparison among models depended materially on
the point values assigned to those intervals.

For each model, we retained the out-of-fold predictions obtained under
the primary training procedure and recomputed uncertain-case MAE using
the lower-endpoint and interval-midpoint mappings. Table
~\ref{tab:mapping_sensitivity} reports the resulting values for the
representative XGBoost, TabPFN, and neural configurations.

\begin{table}[htbp]
    \centering
    \small
    \begin{tabular}{lccc}
        \toprule
        \textbf{Model}
        & \textbf{Lower-endpoint}
        & \textbf{Midpoint}
        & \textbf{Primary-mapping} \\
        \midrule
        XGBoost observed-only
            & 0.433 $\pm$ 0.019 & 0.418 $\pm$ 0.019 & 0.405 $\pm$ 0.020 \\
        XGBoost expert-average
            & 0.182 $\pm$ 0.015 & 0.172 $\pm$ 0.014 & 0.164 $\pm$ 0.014 \\
        TabPFN observed-only
            & 0.396 $\pm$ 0.020 & 0.382 $\pm$ 0.020 & 0.369 $\pm$ 0.020 \\
        TabPFN expert-average
            & 0.178 $\pm$ 0.015 & 0.169 $\pm$ 0.015 & 0.162 $\pm$ 0.015 \\
        Neural observed-only
            & 0.438 $\pm$ 0.014 & 0.424 $\pm$ 0.014 & 0.410 $\pm$ 0.014 \\
        Neural moderate alignment
            & 0.132 $\pm$ 0.010 & 0.126 $\pm$ 0.010 & 0.122 $\pm$ 0.010 \\
        Neural strong alignment
            & 0.045 $\pm$ 0.006 & 0.047 $\pm$ 0.006 & 0.053 $\pm$ 0.007 \\
        \bottomrule
    \end{tabular}
    \caption{\textbf{Sensitivity of uncertain-case MAE to the numerical
    mapping of the expert response categories.}
    Reported values are mean $\pm$ standard deviation across the five
    outer folds. Fitted model predictions are held fixed; only the
    patient-level expert-derived reference values are recomputed under
    the lower-endpoint, interval-midpoint, and primary mappings. The
    analysis therefore assesses sensitivity of uncertain-case evaluation
    rather than sensitivity of model training.}
    \label{tab:mapping_sensitivity}
\end{table}

Across the evaluated models, uncertain-case MAE shifts by only a few
thousandths to a few hundredths across the three mappings, and the
comparison between observed-only and expert-informed training is
unchanged by this choice. In particular, both expert-average baselines
(XGBoost, TabPFN) cut MAE by roughly 0.20--0.25 relative to their
observed-only counterparts under every mapping, and the neural models
show the same pattern at a larger scale: the moderately aligned
configuration reduces MAE by a factor of roughly 3--3.5 relative to the
observed-only neural model, and the strongly aligned configuration by
close to an order of magnitude, again under all three mappings. The one
model-specific detail is direction: MAE decreases monotonically from the
lower-endpoint to the primary (upper-bound) mapping for every model
except the strongly aligned neural configuration, whose predictions are
already concentrated near the low end of the expert-response range and
whose MAE therefore increases slightly (0.045 to 0.053) as the mapping
shifts upward. Thus, although the absolute MAE values vary with the
numerical representation of the expert response intervals, the
principal comparison between observed-only and expert-informed models
remains stable across mappings.

\subsection{Full Neural Sweep}

\begin{figure}[htbp]
    \centering
    \includegraphics[width=0.9\textwidth]{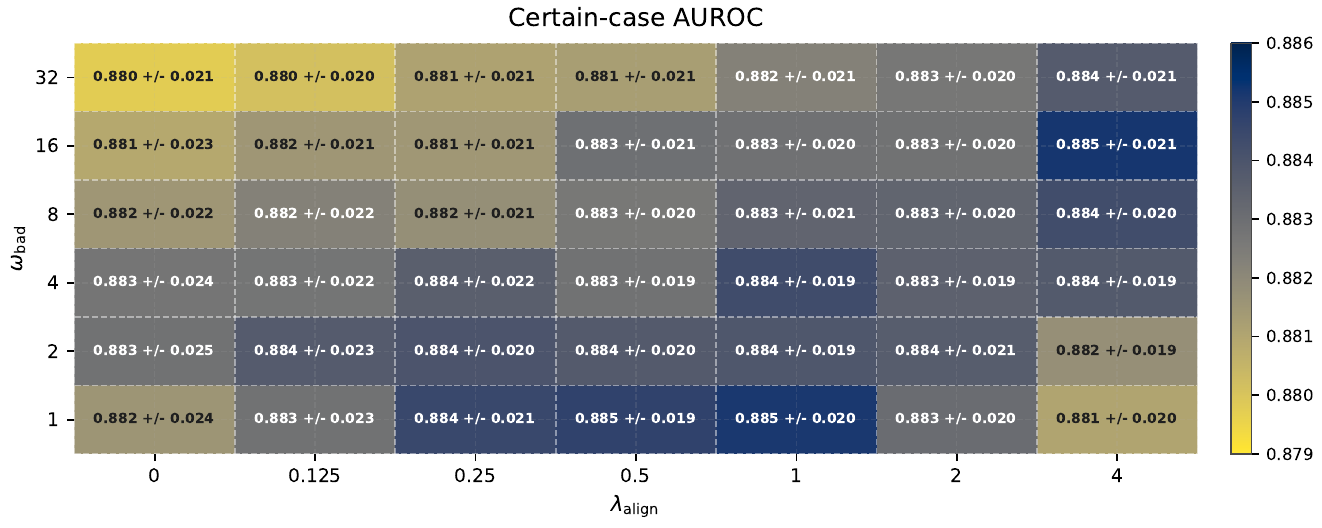}
    \includegraphics[width=0.9\textwidth]{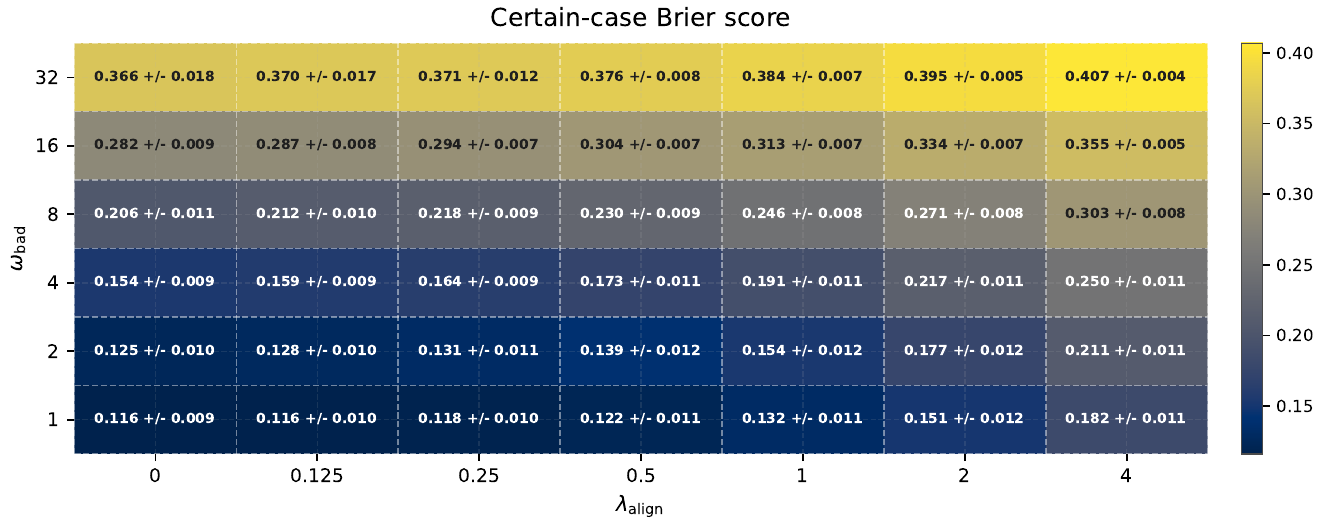}
    \includegraphics[width=0.9\textwidth]{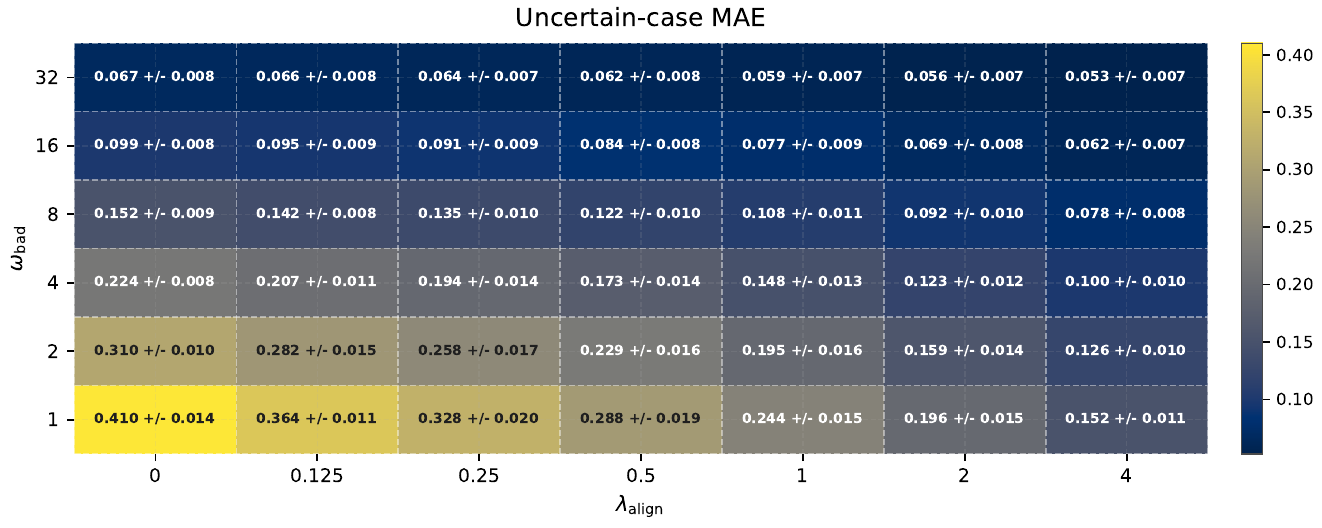}
    \caption{\textbf{Neural sweep summary across $(\omega_{\mathrm{bad}}, \lambda_{\mathrm{align}})$.}
    This figure provides the full numerical view of the neural tradeoff patterns summarized in Figure~\ref{fig:trade-off-scatter-main-paper} and Table~\ref{tab:representative_tradeoff} of the main paper. 
    Each heatmap summarizes one metric over the full 42-model neural sweep. Rows correspond to poor-outcome weights $\omega_{\mathrm{bad}}$ and columns correspond to uncertain-case alignment strengths $\lambda_{\mathrm{align}}$. Top: certain-case AUROC. Middle: certain-case Brier score. Bottom: uncertain-case MAE relative to the expert mean. The sweep shows that the certain--uncertain tradeoff is smooth rather than isolated to a few operating points: AUROC remains comparatively stable, while uncertain-case MAE improves and certain-case Brier score worsens as poor-outcome weighting and uncertain-case alignment increase.}
    \label{fig:nn-heatmaps}
\end{figure}

To complement the aggregate scatter plots, we summarize the full 42-model neural sweep across the two method parameters $(\omega_{\mathrm{bad}}, \lambda_{\mathrm{align}})$. Figure~\ref{fig:nn-heatmaps} reports one heatmap each for certain-case AUROC, certain-case Brier score, and uncertain-case MAE. Together, these heatmaps make the tradeoff surface explicit: AUROC is relatively stable over much of the sweep, whereas Brier score and uncertain-case MAE move strongly and in opposing directions as poor-outcome weighting and uncertain-case alignment increase.

\subsection{Prediction Distributions Across the Full Baseline Family}

\begin{figure}[htbp]
    \centering 
    \includegraphics[width=0.95\textwidth]{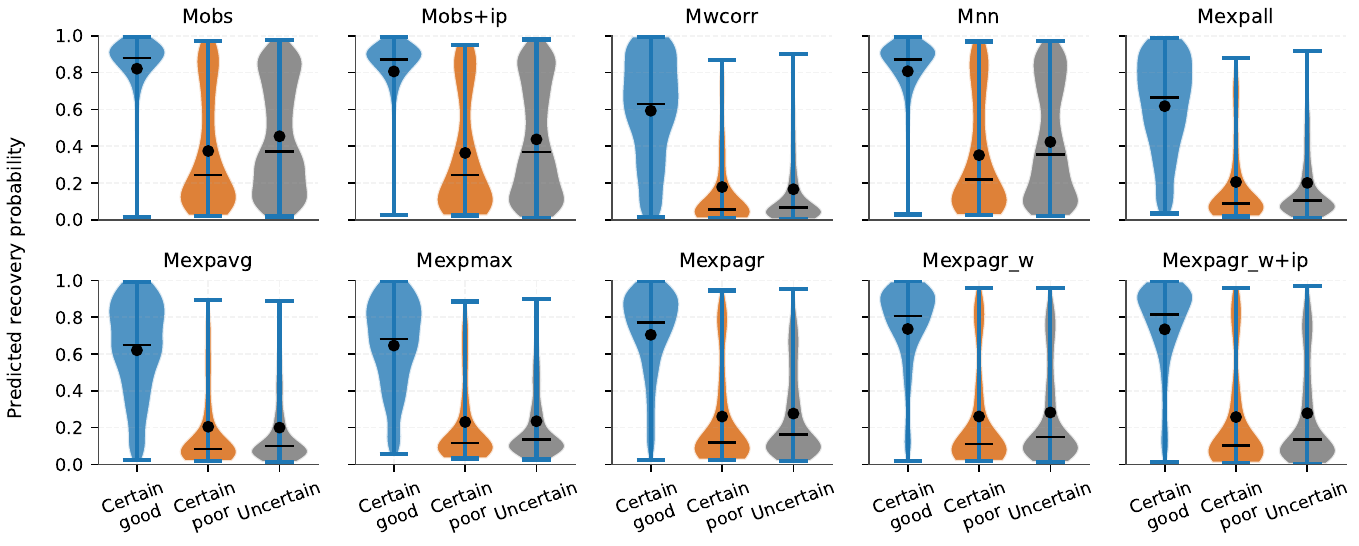} 
    \caption{\textbf{Prediction distributions across XGBoost label-handling baselines.}
    Out-of-fold predicted recovery probabilities are shown for certain good, certain poor, and uncertain cases. The observed-only variant places uncertain cases relatively high on the probability scale, whereas expert-informed variants shift uncertain-case predictions downward. More aggressive uncertain-case incorporation improves uncertain-case agreement on average but tends to reduce separation between certain good cases and the low-probability region.}
    \label{fig:pred-dist-xgboost} 
\end{figure}

\begin{figure}[htbp]
    \centering 
    \includegraphics[width=0.95\textwidth]{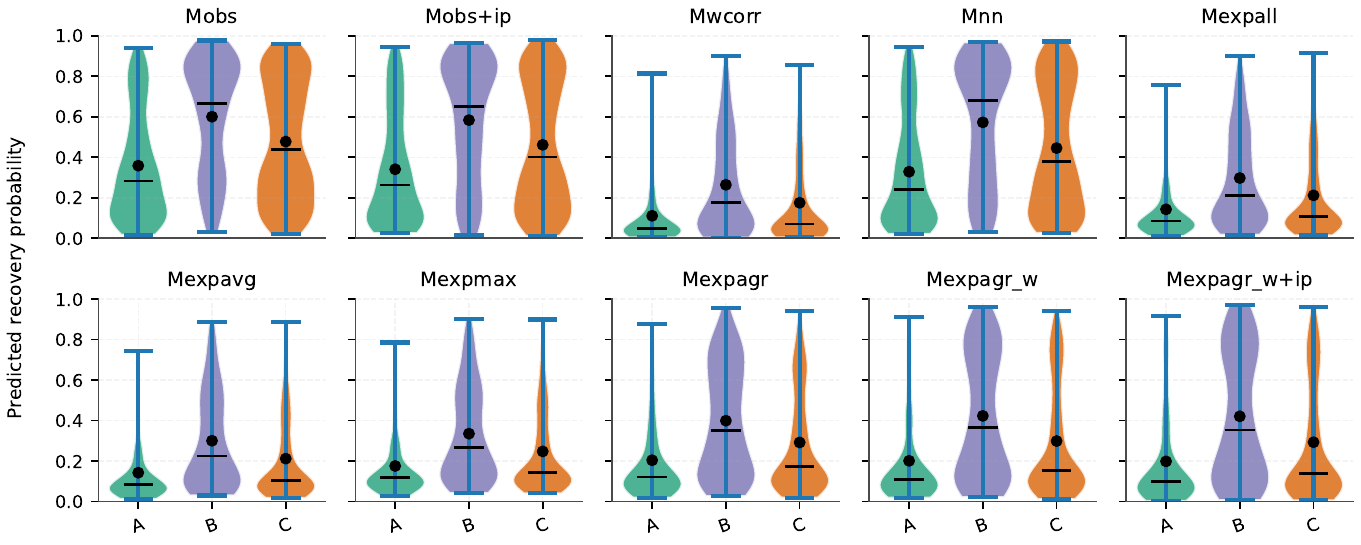} 
    \caption{\textbf{Uncertain-cohort subgroup distributions across XGBoost baselines.}
    Uncertain cases are partitioned into expert-defined groups A--C around the 1\% recovery threshold. Expert-informed XGBoost baselines improve separation of group A from the higher-probability uncertain subgroups, but stronger uncertain-case handling rules also compress group-B predictions downward.}
    \label{fig:pred-dist-uncertain-group-xgboost} 
\end{figure}

\begin{figure}[htbp]
    \centering 
    \includegraphics[width=0.95\textwidth]{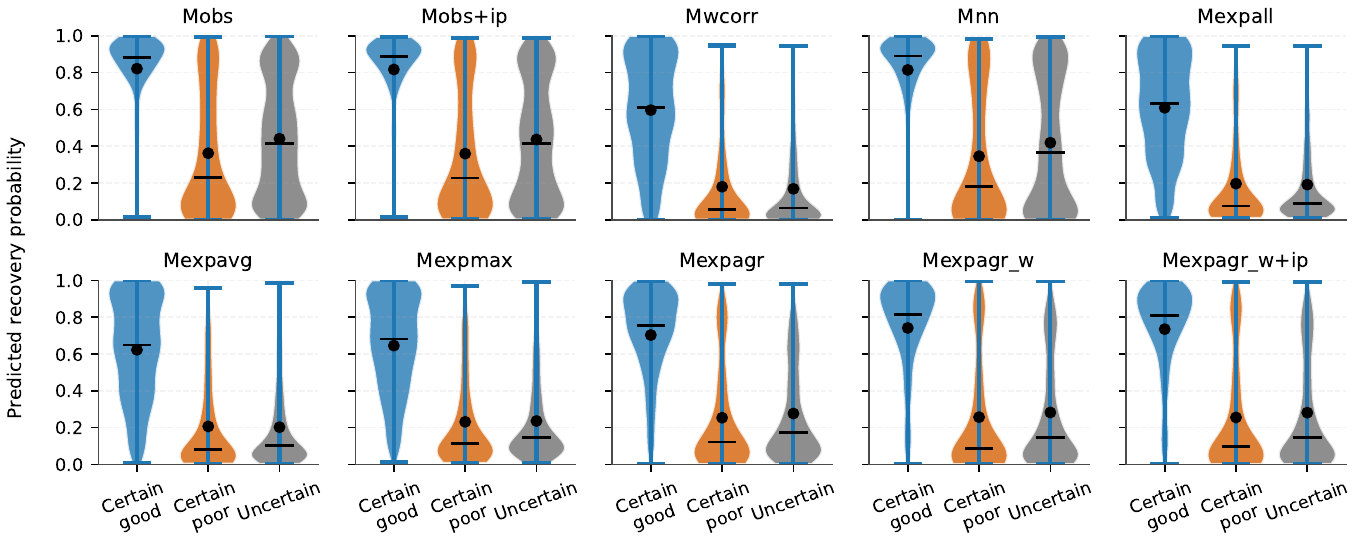} 
    \caption{\textbf{Prediction distributions across random-forest label-handling baselines.}
    Random-forest baselines show the same qualitative pattern as XGBoost: observed-only training places uncertain cases relatively high on the recovery scale, while expert-informed variants improve uncertain-case agreement only by shifting predictions downward and worsening certain-case probability placement.}
    \label{fig:pred-dist-rf} 
\end{figure}

\begin{figure}[htbp]
    \centering 
    \includegraphics[width=0.95\textwidth]{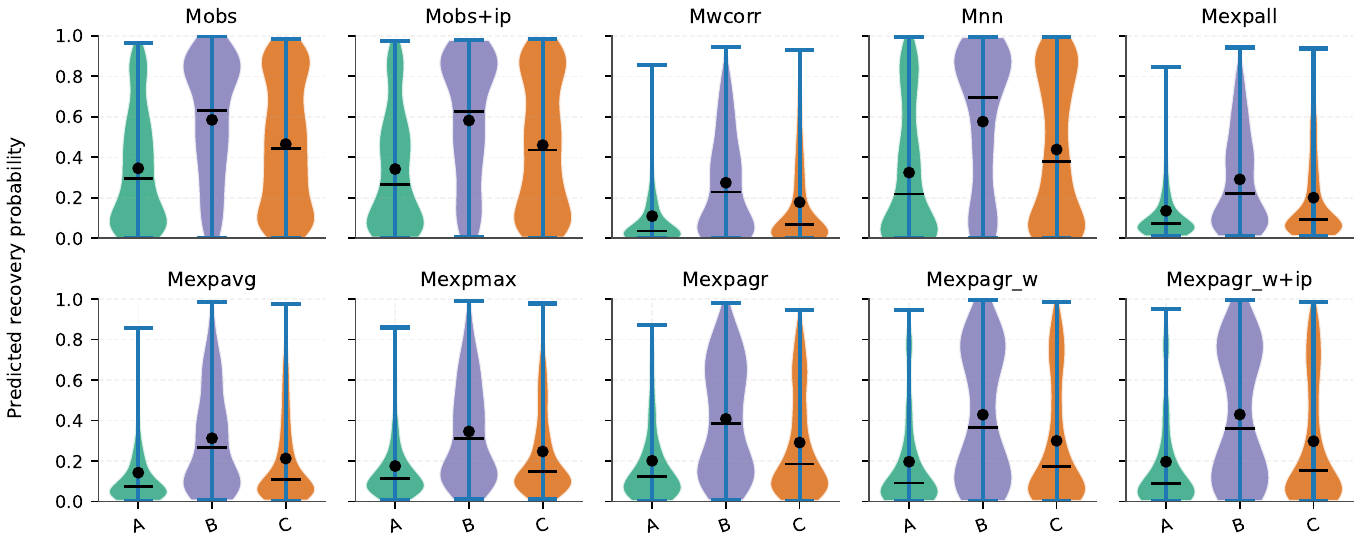} 
      \caption{\textbf{Uncertain-cohort subgroup distributions across random-forest baselines.}
    Groupwise prediction distributions again show that more aggressive uncertain-case handling improves placement for the lowest-probability uncertain patients but risks compressing clinically heterogeneous uncertain subgroups into a uniformly low-probability range.}
    \label{fig:pred-dist-uncertain-group-rf} 
\end{figure}

\begin{figure}[htbp]
    \centering 
    \includegraphics[width=1\textwidth]{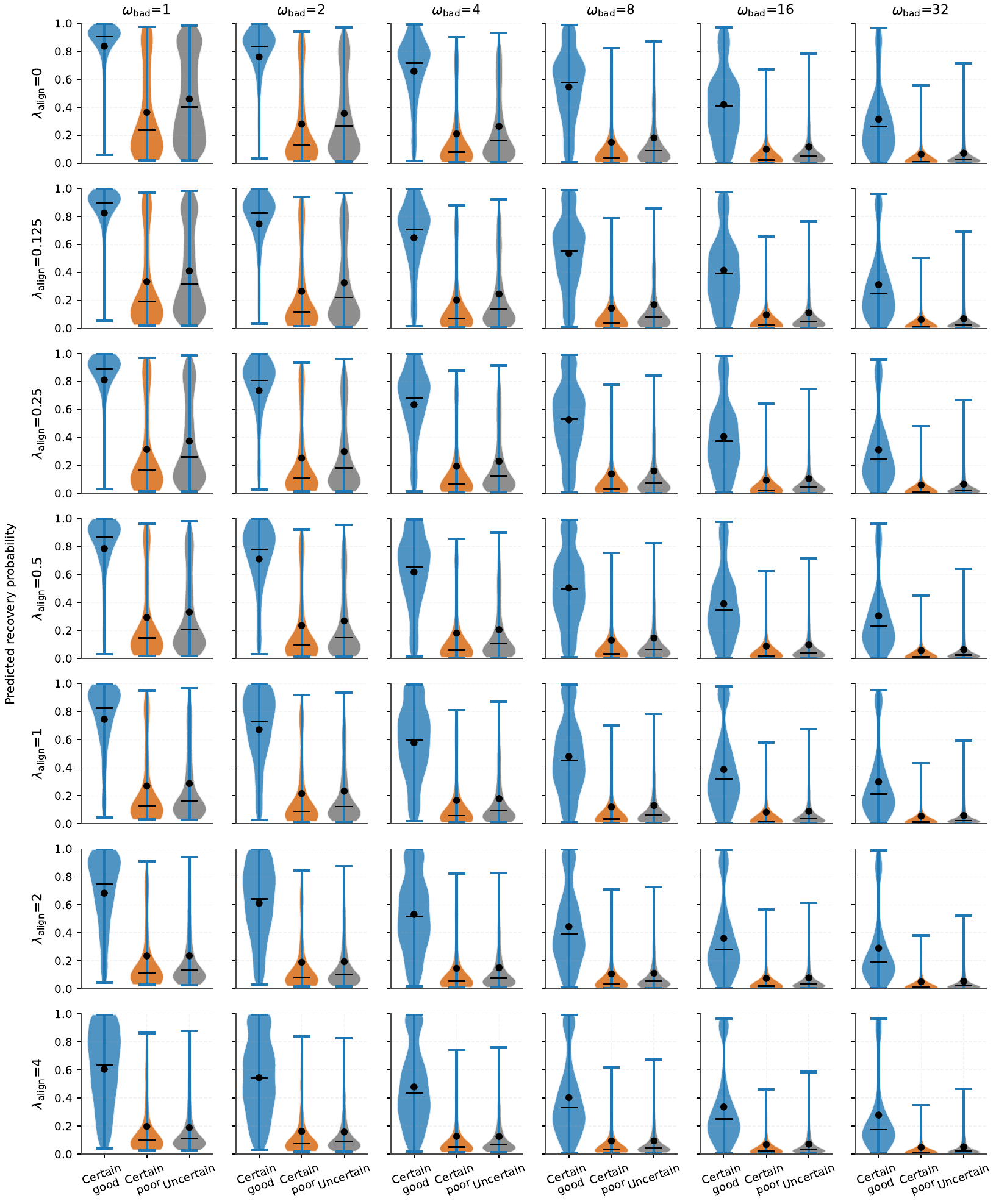} 
    \caption{\textbf{Prediction distributions across the neural operating-point grid.}
    The neural sweep exposes the certain--uncertain tradeoff continuously. As $\omega_{\mathrm{bad}}$ and $\lambda_{\mathrm{align}}$ increase, uncertain-case predictions shift downward and overlap less with certain good cases, but sufficiently strong settings also compress certain poor and uncertain cases together and lower predictions for certain good cases.}
    \label{fig:pred-dist-NN} 
\end{figure}

\begin{figure}[htbp]
    \centering 
    \includegraphics[width=1\textwidth]{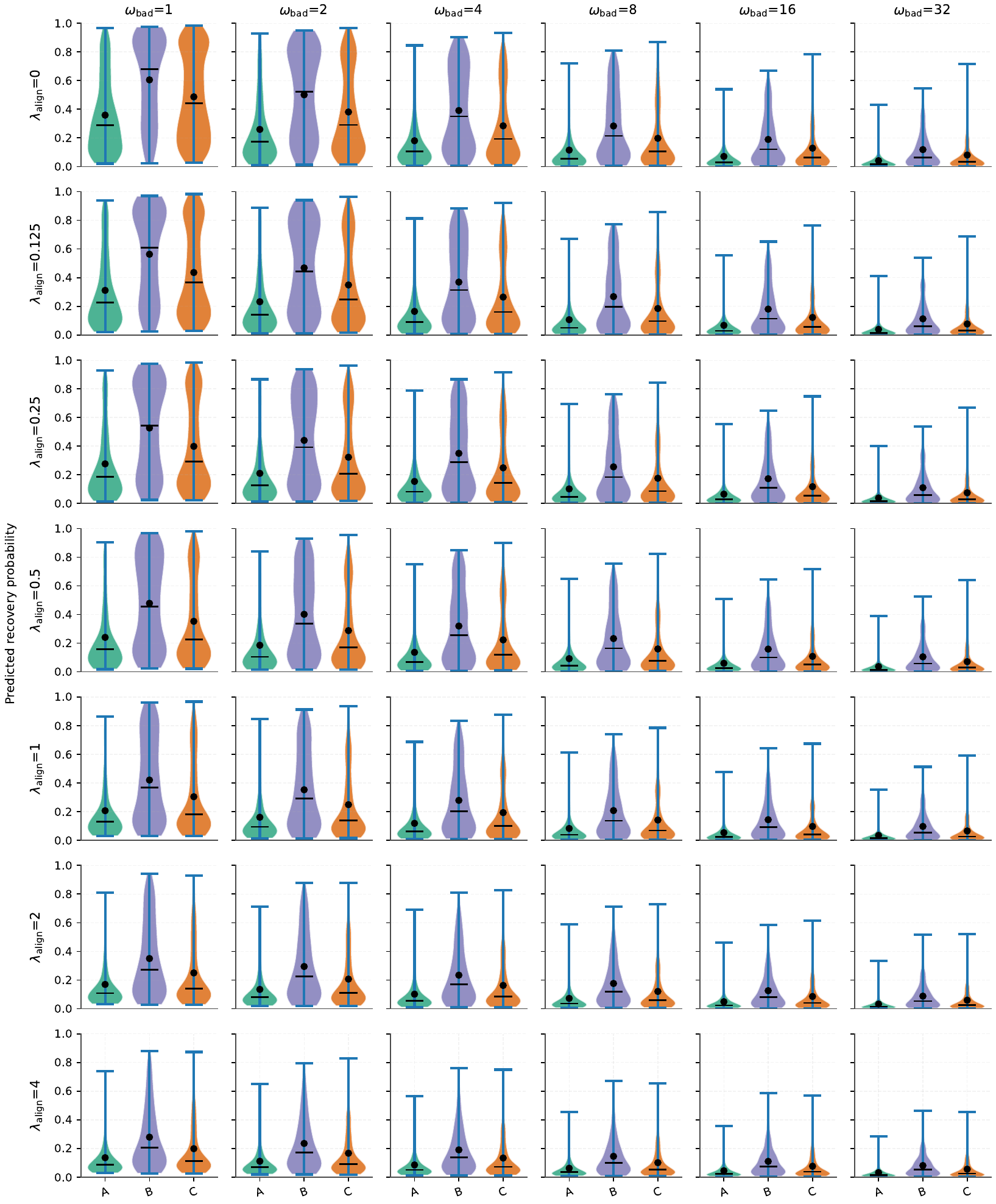} 
    \caption{\textbf{Uncertain-cohort subgroup distributions across the neural operating-point grid.}
    Moderate neural alignment creates an intermediate regime in which group A moves toward the near-zero region while groups B and C remain distinguishable. Stronger alignment further improves average uncertain-case agreement but increasingly compresses all uncertain subgroups downward, including group B.}
    \label{fig:pred-dist-uncertain-group-NN} 
\end{figure}

The main paper uses five representative models to illustrate how the certain--uncertain tradeoff manifests on the prediction scale.
Figures~\ref{fig:pred-dist-xgboost}--\ref{fig:pred-dist-uncertain-group-NN} extend this view to the complete XGBoost and random-forest strategy suites and the neural operating-point grid. TabPFN is included in the aggregate and numerical comparisons, but additional TabPFN distribution panels are omitted to avoid duplicating the same qualitative pattern.

For both XGBoost and random forest, the same general pattern appears across label-handling strategies. Observed-only models assign comparatively high recovery probabilities to uncertain patients and produce substantial overlap between uncertain cases and certain good cases. Strategies that incorporate expert-derived uncertain-case information shift uncertain-case predictions downward and improve average uncertain-case agreement, but increasingly aggressive strategies also compress predictions toward the low-probability region. The neural grid shows this transition most continuously: increasing $\omega_{\mathrm{bad}}$ and especially $\lambda_{\mathrm{align}}$ progressively lowers uncertain-case predictions, but strong settings also pull certain good cases downward, matching the degradation in certain-case Brier score seen in the aggregate results.

\newpage
\subsection{Groupwise Uncertain-Case Results}
\label{app:groupwise_uncertain_results}

To further interpret model behavior within the uncertain cohort, we summarize predictions separately for the three expert-defined uncertain-case groups defined in Appendix~\ref{sec:expert-group-abc} and summarized in Figure~\ref{fig:expert-score-dist}: group A (all experts $\leq 1\%$), group B (all experts $>1\%$), and group C (expert disagreement across the 1\% threshold). This analysis is intended to clarify how the neural sweep reshapes uncertain-case predictions relative to the clinically important 1\% recovery threshold, rather than to redefine the main accuracy--alignment tradeoff.

\begin{figure}[t]
    \centering
    \includegraphics[width=\textwidth]{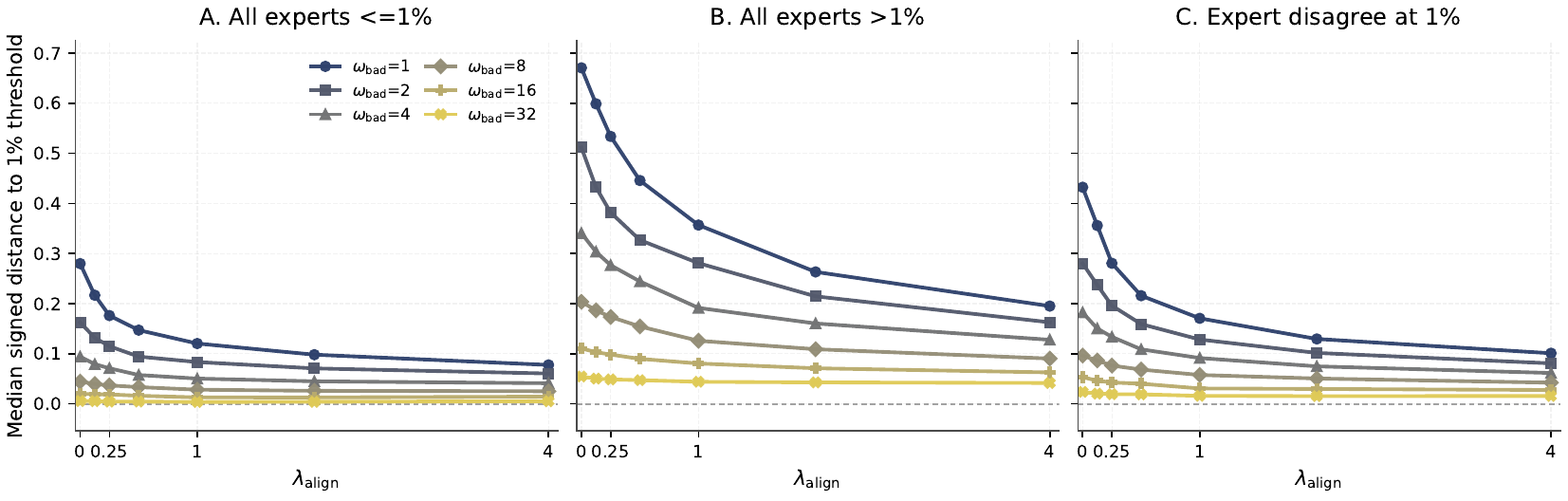}
    \caption{\textbf{Threshold-relative movement of uncertain-case subgroup predictions across the neural sweep.}
    Each panel corresponds to one expert-defined uncertain subgroup: group A (all experts $\leq 1\%$), group B (all experts $>1\%$), and group C (expert disagreement across the 1\% threshold). The x-axis is uncertain-case alignment strength $\lambda_{\mathrm{align}}$, and each line corresponds to one poor-outcome weight $\omega_{\mathrm{bad}}$. For each subgroup and model setting, the y-axis shows the median signed distance from the 1\% recovery threshold, defined as $\mathrm{median}_{i \in g}(\hat y_i - 0.01)$, where $\hat y_i$ is the predicted probability of recovery for uncertain patient $i$. Zero denotes the threshold itself; positive values indicate subgroup median predicted probability above 1\%. Across groups, stronger poor-outcome weighting and uncertain-case alignment move predictions downward toward the threshold while largely preserving subgroup ordering.}
    \label{fig:nn_uncertain_group_threshold_movement}
\end{figure}

Figure~\ref{fig:nn_uncertain_group_threshold_movement} shows subgroup-level threshold-relative movement across the full neural sweep. For each model setting $(\omega_{\mathrm{bad}}, \lambda_{\mathrm{align}})$ and each uncertain-case subgroup, we first compute a patient-level signed distance to the 1\% threshold,
\[
d_i = \hat y_i - 0.01,
\]
where $\hat y_i$ is the predicted probability of recovery for uncertain patient $i$. The plotted y-axis value is then the median of this quantity within the subgroup,
\[
\mathrm{median}_{i \in g}(\hat y_i - 0.01),
\]
where $g \in \{A,B,C\}$ denotes the expert-defined subgroup. Positive values therefore indicate that the subgroup median prediction lies above 1\%, negative values indicate that it lies below 1\%, and zero corresponds exactly to the 1\% threshold.

Several patterns are apparent. First, increasing either the alignment strength $\lambda_{\mathrm{align}}$ or the poor-outcome weight $\omega_{\mathrm{bad}}$ generally shifts predictions downward toward the 1\% threshold in all three groups. Second, the relative ordering of the expert-defined subgroups is largely preserved across the sweep: group B remains highest, group C remains intermediate, and group A remains lowest. Thus, stronger alignment does not mainly operate by reversing subgroup ordering. Instead, it compresses all uncertain subgroups toward lower predicted recovery probabilities.

This decomposition also helps interpret the main-paper observation that aggressive alignment can overcompress uncertain-case predictions. Group B is especially informative in this regard, because all experts judged these patients to have probability of recovery exceeding 1\%. As alignment becomes stronger, group-B predictions move substantially closer to the 1\% threshold, even though the subgroup median generally remains above it. The practical concern is therefore not simply whether uncertain-case error decreases on average, but whether increasingly aggressive alignment compresses clinically heterogeneous uncertain patients toward the near-zero region more than is clinically warranted.

\end{document}